\documentclass[twoside,11pt]{article}
\usepackage{jmlr2e}
\usepackage{tabularx}
\usepackage{url}
\usepackage{booktabs}
\usepackage{color,epsf,psfrag}
\usepackage{algorithm, algorithmic}
\usepackage{amsmath,amsbsy,amsfonts,amssymb,dsfont,units,url}
\usepackage{tikz}
\usetikzlibrary{calc,shapes}

\def\fighome{.}
\def\Rbb{\mathbb{R}}
\newcommand{\llvert}{\left\vert\vphantom{\frac{1}{1}}\right.}

\newcommand{\Pvalue}{\mathsf{P_{\text{val}}}}
\newcommand{\Pvaluem}{\mathsf{\mathbf{P}_{\text{val}}}}
\usepackage{float}
\usepackage{cases}
\usepackage{enumitem}

\DeclareMathOperator{\Ber}{Ber}
\DeclareMathOperator{\topic}{Top}
\DeclareMathOperator{\community}{Com}

\def\tha{{\mbox{\tiny th}}}

\def\nx{n_{\mbox{\tiny\itshape \!X}}}

\newcommand\R{\mathbb{R}}

\DeclareMathOperator{\diag}{diag}
\DeclareMathOperator{\degree}{Degree}

\DeclareMathOperator{\Pairs}{Pairs}

\newcommand\Dir{\operatorname{Dir}}

\newcommand\Poi{\operatorname{Poi}}
\newcommand{\bp}{\begin{psfrags}}
\newcommand{\ep}{\end{psfrags}}

\newcommand{\bprf}{\begin{myproof}}
\newcommand{\eprf}{\end{myproof}}

\newenvironment{myproof}{\noindent{\em Proof:} \hspace*{1em}}{
    \hspace*{\fill} $\Box$ }
\newenvironment{proof_of}[1]{\noindent {\em Proof of #1: }}{\hspace*{\fill} $\Box$ }

\def\viz{{viz.,\ \/}}
\def\Ebb{{\mathbb E}}

\def\beq{\begin{equation}}
\def\eeq{\end{equation}\noindent}
\def\beqn{\begin{eqnarray}}
\def\eeqn{\end{eqnarray} \noindent}
\def\beqnn{\begin{eqnarray*}}
\def\eeqnn{\end{eqnarray*}  \noindent}
\def\bcase{\begin{numcases}}
\def\ecase{\end{numcases}   \noindent}

\jmlrheading{ }{2014}{ }{ }{ }{Huang et al.}
\ShortHeadings{Online Tensor Methods for Learning Latent Variable Models}{Huang et al.}
\firstpageno{1}
\begin{document}

\title{Online Tensor Methods for Learning Latent Variable Models}

\author{\name Furong Huang \email furongh@uci.edu \\
       \addr Electrical Engineering and Computer Science Dept.\\
       University of California, Irvine\\
       Irvine, USA 92697, USA
       \AND
       \name U. N. Niranjan \email un.niranjan@uci.edu \\
       \addr Electrical Engineering and Computer Science Dept.\\
       University of California, Irvine\\
       Irvine, USA 92697, USA
       \AND
       \name Mohammad Umar Hakeem \email mhakeem@uci.edu \\
       \addr Electrical Engineering and Computer Science Dept.\\
       University of California, Irvine\\
       Irvine, USA 92697, USA
       \AND
       \name Animashree Anandkumar \email a.anandkumar@uci.edu \\
       \addr Electrical Engineering and Computer Science Dept.\\
       University of California, Irvine\\
       Irvine, USA 92697, USA}

\editor{Charles Sutton}
\maketitle
\begin{abstract}
We introduce an online tensor decomposition based approach for two latent variable modeling problems namely, (1) community detection,  in which we learn the latent communities that the social actors in social networks belong to, and (2) topic modeling, in which we infer hidden topics of text articles. We consider decomposition of moment tensors using stochastic gradient descent. We conduct   optimization of multilinear operations  in SGD and  avoid directly forming the tensors, to save computational and storage costs. 
We present optimized algorithm in two platforms.  Our GPU-based implementation exploits the parallelism of SIMD architectures to allow for maximum speed-up by a careful optimization of storage and data transfer, whereas our CPU-based implementation uses efficient sparse matrix computations and is suitable for large sparse datasets. For the community detection problem, we demonstrate accuracy and computational efficiency on Facebook, Yelp and DBLP datasets, and for the topic modeling problem, we also demonstrate good performance on the New York Times dataset. 
We compare our results to the state-of-the-art algorithms such as the variational method, and report a gain of accuracy and a gain of several orders of magnitude in the execution time.
\end{abstract}

\section{Introduction}

The spectral or  moment-based approach involves decomposition of   certain empirical moment tensors, estimated from observed data to obtain  the parameters of the proposed probabilistic model. Unsupervised learning for a wide range of latent variable models can be carried out efficiently via tensor-based techniques with low sample and computational complexities~\citep{AGHKT12}. In contrast, usual methods employed in practice such as expectation maximization (EM) and variational Bayes do not have such consistency guarantees. While the previous works~\citep{AnandkumarEtal:community12COLT} focused on theoretical guarantees, in this paper, we focus on the implementation of the tensor methods, study its performance on several datasets.

\subsection{Summary of Contributions}

We consider two problems: (1) community detection (wherein we compute the  decomposition of a tensor which relates to the count of $3$-stars in a graph) and (2) topic modeling (wherein we consider the tensor related to co-occurrence of triplets of words in documents); decomposition of the these tensors allows us to learn the hidden communities and topics from observed data.

\paragraph{Community detection: } We recover hidden communities in several real datasets with high accuracy. When ground-truth communities are available, we propose a new error score based on the hypothesis testing methodology involving $p$-values and false discovery rates~\citep{strimmer2008fdrtool} to validate our results. 
The use of $p$-values eliminates the need to carefully tune the number of communities output by our algorithm, and hence, we obtain a flexible trade-off between the fraction of communities recovered and their estimation accuracy.  
We find that our method has very good accuracy on a range of network datasets: Facebook, Yelp and DBLP. We summarize the datasets used in this paper in Table~\ref{tab:data_info}. To get an idea of our running times, let us consider the larger DBLP collaborative dataset for a moment. It consists of $16$ million edges, one million nodes and $250$ communities. We obtain an error of $10\%$ and the method runs in about two minutes, excluding the $80$ minutes taken to read the edge data from files stored on the hard disk and converting it to sparse matrix format.

Compared to the state-of-the-art method for learning MMSB models using the stochastic variational inference algorithm of~\citep{gopalan2012scalable}, we obtain several orders of magnitude speed-up in the running time on multiple real datasets. This is because our method consists of efficient matrix operations which are \emph{embarrassingly parallel}. Matrix operations are   carried out in the sparse format which is efficient especially for social network settings involving large sparse graphs. Moreover, our code is flexible to run on a range of graphs such as directed, undirected and bipartite graphs, while the code of~\citep{gopalan2012scalable} is designed for homophilic networks, and cannot  handle bipartite graphs in its present format. Note that bipartite networks occur in the recommendation setting such as the Yelp dataset. Additionally, the variational implementation in~\citep{gopalan2012scalable} assumes a homogeneous connectivity model, where any pair of communities connect with the same probability and the probability of intra-community connectivity is also fixed. Our framework does not suffer from this restriction. We also provide arguments to show that the Normalized Mutual Information (NMI) and other scores, previously used for evaluating the recovery of overlapping community, can underestimate the errors.

\paragraph{Topic modeling: }We also employ the tensor method for topic-modeling, and there are many similarities between the topic and community settings. For instance, each document has  multiple topics, while in the network setting, each node has membership in multiple communities. The words in a document are generated based on the latent topics in the document, and similarly, edges are generated based on the community memberships of the node pairs. The tensor method is even faster for topic modeling, since the word vocabulary size is typically much smaller than the size of real-world networks.   We learn interesting hidden topics in New York Times corpus from UCI bag-of-words dataset\footnote{\url{https://archive.ics.uci.edu/ml/datasets/Bag+of+Words}} with around $100,000$ words and $300,000$ documents in about two minutes. We present the important words for recovered topics, as well as interpret ``bridging'' words, which occur in many topics. 


\paragraph{Implementations: }We present two implementations, \viz a GPU-based implementation which exploits the parallelism of SIMD architectures and a CPU-based implementation for larger datasets, where the GPU memory does not suffice. We discuss various aspects   involved such as implicit manipulation of tensors since explicitly forming tensors would be unwieldy for large networks,   optimizing for communication bottlenecks in a parallel deployment, the need for sparse matrix and vector operations since real world networks tend to be sparse, and a careful statistical approach to validating the results, when ground truth is available.

\subsection{Related work}
This paper builds on the recent works of Anandkumar et al~\citep{AGHKT12,AnandkumarEtal:community12COLT} which establishes the correctness of tensor-based approaches for learning MMSB~\citep{ABFX08} models and other latent variable models.
While, the earlier works provided a theoretical analysis of the method, the current paper considers a careful implementation of the method. Moreover, there are a number of algorithmic improvements in this paper. For instance, while \citep{AGHKT12,AnandkumarEtal:community12COLT} consider tensor power iterations, based on batch data and deflations performed serially, here, we adopt a  stochastic gradient descent approach for tensor decomposition, which provides the flexibility to trade-off  sub-sampling with accuracy. Moreover, we use randomized methods for dimensionality reduction in the preprocessing stage of our method which enables us to  scale our method to graphs with millions of nodes.

There are other known methods for learning the stochastic block model based on techniques such as spectral clustering~\citep{McSherry01} and convex optimization~\citep{chen2012clustering}. However, these methods are not applicable for learning overlapping communities.
We note that learning
the mixed membership model can be reduced to a matrix factorization problem~\citep{Zhang:2012:OCD:2339530.2339629}. While collaborative filtering techniques  such as~\citep{mnih2007probabilistic,salakhutdinov2008bayesian}
focus on matrix factorization and the prediction accuracy of recommendations on an unseen test set, we recover the underlying latent communities, which helps with the interpretability and the statistical model can be employed for other tasks.

Although there have been other fast implementations for community detection before~\citep{soman2011fast,lancichinetti2009community}, these methods are not statistical and do not yield descriptive statistics such as bridging nodes~\citep{nepusz2008fuzzy}, and cannot perform predictive tasks such as link classification which are the main strengths of the MMSB model. With the implementation of our tensor-based approach, we record huge speed-ups compared to existing approaches for learning the MMSB model.

To the best of our knowledge, while stochastic methods for matrix decomposition have been considered earlier~\citep{oja1985stochastic,6483308}, this is the first work incorporating stochastic optimization for tensor decomposition, and paves the way for further investigation on many theoretical and practical issues.
We also note that we never explicitly form or store the subgraph count tensor, of size $O(n^3)$ where $n$ is the number of nodes, in our implementation, but directly manipulate the neighborhood vectors to obtain tensor decompositions through stochastic updates. This is a crucial departure from other works on tensor decompositions on GPUs~\citep{ballard2011efficiently,schatz2013exploiting}, where the tensor needs to be stored and manipulated directly.

\section{Tensor Forms for Topic and Community  Models}
\label{sec:sysmodel}
In this section, we briefly recap the topic and community models, as well as the tensor forms for their exact moments, derived in~\citep{AGHKT12,AnandkumarEtal:community12COLT}.

\subsection{Topic Modeling}
In  topic modeling, a document is viewed as a bag of words. Each document has a latent set of topics, and $h=(h_1,h_2,\ldots,h_k)$ represents the proportions of $k$ topics in a given document.  Given the topics $h$, the words are independently drawn and are exchangeable, and hence, the term ``bag of words'' model. We represent the
words in the document by $d$-dimensional random vectors $x_1, x_2, \ldots x_l \in \mathbb{R}^d$, where $x_i$ are coordinate basis vectors in $\mathbb{R}^d$ and $d$ is the size of the word vocabulary. Conditioned on $h$, the words in a document satisfy $\Ebb[x_i|h]=\mu h$, where $\mu : = [\mu_1,\ldots,\mu_k]$ is the topic-word matrix. And thus $\mu_j $ is the topic vector satisfying $\mu_j = \Pr\left(x_i \vert h_j\right)$,  $\forall j\in[k]$.
Under the Latent Dirichlet Allocation (LDA) topic model~\citep{blei2012probabilistic}, $h$ is drawn from a Dirichlet distribution with concentration parameter vector $\alpha = [\alpha_1,\ldots,\alpha_k]$. In other words, for each document $u$,  $h_u \stackrel{iid}{\sim}\Dir(\alpha),\ \forall u\in [n]$ with parameter vector $\alpha \in \R_{+}^k$. We define the Dirichlet concentration (mixing) parameter
\[
\alpha_0:=\sum_{i\in[k]}{\alpha_i}.
\]
  The Dirichlet distribution allows us to specify the extent of overlap among the topics by controlling for sparsity in topic density function. A larger $\alpha_0$ results in   more overlapped (mixed) topics. A special case of $\alpha_0=0$ is the single topic model.

Due to exchangeability, the order of the words does not matter, and it suffices to consider the frequency vector for each document, which counts the number of occurrences of each word in a document. Let $c_t:= (c_{1,t}, c_{2,t},\ldots, c_{d,t})\in\Rbb^{d}$ denote the frequency vector for $t^{\tha}$ document, and let $n$ be the number of documents.

We consider the first three order empirical moments,  given by
\begin{align}
\label{eq:1moment_topic}
M_1^{\topic} &: = \frac{1}{n} \sum\limits_{t=1}^{n} c_t\\
\label{eq:2moment_topic}
M_2^{\topic} &:=
\frac{\alpha_0+1}{n} \sum\limits_{t=1}^{n}{\left(c_t\otimes c_t - \diag\left(c_t\right)\right)} - {\alpha_0}M_1^{\topic}\otimes M_1^{\topic}\\
\label{eq:3moment_topic}
M_3^{\topic} & :=
 \frac{(\alpha_0+1)(\alpha_0+2)}{2n}\sum\limits_{t=1}^{n}\left[  c_t\otimes c_t\otimes c_t - \sum\limits_{i=1}^{d} \sum\limits_{j=1}^{d}c_{i,t}c_{j,t}(e_i\otimes e_i\otimes e_j) - \sum\limits_{i=1}^{d} \sum\limits_{j=1}^{d}c_{i,t}c_{j,t}(e_i\otimes e_j\otimes e_i) \right. \nonumber \\
& \left. - \sum\limits_{i=1}^{d} \sum\limits_{j=1}^{d} c_{i,t}c_{j,t}(e_i\otimes e_j\otimes e_j) + 2 \sum\limits_{i=1}^{d} c_{i,t}(e_i\otimes e_i\otimes e_i) \right] \nonumber\\
& -\frac{\alpha_0(\alpha_0+1)}{2n} \sum\limits_{t=1}^{n}\left(   \sum\limits_{i=1}^{d} c_{i,t}(e_i\otimes e_i\otimes M_1^{\topic}) +   \sum\limits_{i=1}^{d} c_{i,t}(e_i \otimes M_1^{\topic}\otimes e_i)     +   \sum\limits_{i=1}^{d} c_{i,t}(M_1^{\topic} \otimes e_i \otimes e_i)    \right) \nonumber\\
& + {\alpha_0^2}M_1^{\topic} \otimes M_1^{\topic} \otimes M_1^{\topic}.
\end{align}
We recall Theorem 3.5 of~\citep{AGHKT12}:
\begin{lemma}\label{lemma:topic}
The exact moments can be factorized as
\begin{align}
\label{eq:1Emoment_topic}
\Ebb[M_1^{\topic}] & = \sum\limits_{i=1}^{k} \frac{\alpha_i}{\alpha_0}\mu_i\\
\label{eq:2Emoment_topic}
\Ebb[M_2^{\topic}] & = \sum\limits_{i=1}^{k} \frac{\alpha_i}{\alpha_0}\mu_i \otimes \mu_i\\
\label{eq:3Emoment_topic}
\Ebb[M_3^{\topic}] & =\sum\limits_{i=1}^{k}  \frac{\alpha_i}{\alpha_0}\mu_i\otimes \mu_i \otimes \mu_i.
\end{align}

where $\mu = [\mu_1,\ldots,\mu_k]$ and $\mu_i = \Pr\left(x_t \vert h=i\right)$, $\forall t\in[l]$. In other words, $\mu$ is the topic-word matrix.
\end{lemma}

From the Lemma~\ref{lemma:topic}, we observe that the first three moments of a LDA topic model have a simple form involving the topic-word matrix $\mu$ and Dirichlet parameters $\alpha_i$. In~\citep{AGHKT12}, it is shown that these parameters  can be recovered under a weak non-degeneracy assumption. We will employ tensor decomposition techniques to learn the parameters.

\subsection{Mixed Membership Model}
In the mixed membership stochastic block model (MMSB), introduced by~\citep{ABFX08}, the edges in a social network are related to the hidden communities of the nodes.
A batch tensor decomposition technique for learning MMSB was derived in~\citep{AnandkumarEtal:community12COLT}.

Let $n$ denote the number of nodes, $k$ the number of communities and $G\in \mathbb{R}^{n \times n}$ the adjacency matrix of the graph.  Each node $i\in [n]$ has an associated community membership vector $\pi_i \in \Rbb^k$, which is a latent variable, and the vectors are contained in a simplex, i.e., \[\sum_{i\in [k]} \pi_u(i)=1, \ \forall u\in [n]\]
where the notation $[n]$ denotes the set $\{ 1, \ldots, n \}$.  Membership vectors are sampled from the Dirichlet distribution $\pi _u \stackrel{iid}{\sim} \Dir(\alpha), \ \forall u\in [n]$ with parameter vector $\alpha \in \R_{+}^k$ where $\alpha_0:=\sum_{i\in[k]}{\alpha_i}$.
As in the topic modeling setting, the Dirichlet distribution allows us to specify the extent of overlap among the communities by controlling for sparsity in community membership vectors. A larger $\alpha_0$ results in   more overlapped (mixed) memberships. A special case of $\alpha_0=0$ is the stochastic block model~\citep{AnandkumarEtal:community12COLT}.

The \emph{community connectivity matrix} is denoted by $P\in [0,1]^{k \times k}$ where $P(a,b)$ measures the connectivity between communities $a$ and $b$, $\forall a,b \in [k]$.  We model the adjacency matrix entries as either of the two settings given below:

\paragraph{Bernoulli model: }This models a network with unweighted edges. It is used for Facebook and DBLP datasets in Section~\ref{sec:results} in our experiments.
\[
G_{ij} \stackrel{iid}{\sim} \Ber (\pi_i ^\top P \pi_j),\, \  \forall i,j\in[n] .
\]

\paragraph{Poisson model~\citep{karrer2011stochastic}: }This models  a network with weighted edges. It is used for the Yelp dataset in Section~\ref{sec:results} to incorporate the review ratings. \[
G_{ij} \stackrel{iid}{\sim} \Poi (\pi_i ^\top P \pi_j),\,\ \forall i,j\in[n].\]

The tensor decomposition approach involves up to third order moments, computed from the observed network. In order to compute the moments, we  partition the  nodes randomly into sets $X,A,B,C$. Let $F_A := \Pi_A^\top P^\top$, $F_B := \Pi_B^\top P^\top$, $F_C := \Pi_C^\top P^\top$ (where $P$ is the community connectivity matrix and $\Pi$ is the membership matrix) and $\hat{\alpha}:=\left( \frac{\alpha_1}{\alpha_0},\ldots,\frac{\alpha_k}{\alpha_0} \right)$ denote the normalized Dirichlet concentration parameter.  We define pairs over $Y_1$ and $Y_2$ as $\Pairs(Y_1,Y_2): = G_{X,Y_1}^\top \otimes G_{X,Y_2}^\top$.  Define the following matrices
\begin{align}\label{eq:transitionMat}
Z_B & := \Pairs\left(A,C\right) \left(\Pairs\left(B,C\right)\right)^\dag,\\
Z_C & := \Pairs\left(A,B\right) \left(\Pairs\left(C,B\right)\right)^\dag.
\end{align}

We consider the first three empirical moments, given by
\begin{align}
\label{eq:1moment_graph}
{M_{1}}^{\community} & : = \frac{1}{\nx}\sum\limits_{x\in X} G_{x,A}^\top \\
\label{eq:2moment_graph}
{M_2}^{\community} &: = \frac{\alpha_0 + 1}{\nx} \sum\limits_{x\in X} Z_C G_{x,C}^\top G_{x,B} Z_B^\top -  {\alpha_0} \left({M_{1}}^{\community} {{M_{1}}^{\community}}^\top \right)\\
\label{eq:3moment_graph}
{M_3}^{\community}  & :=  \frac{(\alpha_0 + 1)(\alpha_0 + 2)}{2\nx}\sum_{x\in X}\left[G^\top_{x,A}\otimes Z_B G^\top_{x,B}\otimes Z_C G^\top_{x,C}\right] + \alpha_0^2 {M_{1}}^{\community}  \otimes {M_{1}}^{\community}  \otimes {M_{1}}^{\community}  \nonumber \\
& - \frac{\alpha_0 (\alpha_0 + 1)}{2 \nx} \sum_{x \in X}\left [ G_{x,A}^\top \otimes Z_B G_{x,B}^\top \otimes {M_{1}}^{\community}  +  G_{x,A}^\top  \otimes {M_{1}}^{\community}  \otimes Z_C G_{x,C}^\top +{M_{1}}^{\community} \otimes Z_B G_{x,B}^\top \otimes Z_C G_{x,C}^\top \right]
\end{align}

We now recap Proposition 2.2 of~\citep{AnandkumarEtal:community12} which provides the form of these moments under expectation.

\begin{lemma}
The exact moments can be factorized as
\begin{align}
\label{eqn:single}
\mathbb{E} [{M_1}^{\community} | \Pi_A, \Pi_B, \Pi_C] & :=  \sum_{i\in [k]} \hat{\alpha}_i (F_A)_i\\
\label{eqn:pair}
\mathbb{E} [{M_2}^{\community} | \Pi_A, \Pi_B, \Pi_C] & :=  \sum_{i\in [k]} \hat{\alpha}_i (F_A)_i \otimes (F_A)_i \\
\label{eqn:triples}
\mathbb{E} [{M_3}^{\community} | \Pi_A, \Pi_B, \Pi_C] & := \sum_{i\in [k]} \hat{\alpha}_i (F_A)_i \otimes (F_A)_i \otimes (F_A)_i
\end{align}
where $\otimes$ denotes the {\em Kronecker product} and $(F_A)_i$ corresponds to the $i^{th}$ column of $F_A$.
\end{lemma}

We observe that the moment forms above for the MMSB model have a similar form as the moments of the topic model in the previous section. Thus, we can employ a unified framework for both topic and community modeling  involving decomposition of the third order moment tensors $M_3^{\topic}$ and $M_3^{\community}$. Second order moments $M_2^{\topic}$ and $M_2^{\community}$ are used for \emph{preprocessing} of the data (i.e., whitening, which is introduced in detail in Section~\ref{sec:DRandWhite}). For the sake of the simplicity of the notation, in the rest of the paper, we will use $M_2$ to denote empirical second order moments for both $M_2^{\topic}$ in topic modeling setting, and $M_2^{\community}$ in the mixed membership model setting. Similarly, we will use $M_3$ to denote empirical third order moments for both $M_3^{\topic}$ and $M_3^{\community}$.

\section{Learning using Third Order Moment}

Our learning algorithm uses up to the third-order moment to estimate the topic word matrix $\mu$ or the community membership matrix $\Pi$. First, we obtain co-occurrence of triplet words or subgraph counts (implicitly).
 Then, we perform preprocessing using second order moment $M_2$. Then we perform tensor decomposition efficiently using {\em stochastic gradient descent}~\citep{kushner2003stochastic} on $M_3$. We note that, in our implementation of the algorithm on the Graphics Processing Unit (GPU), linear algebraic operations are extremely fast.
 We also implement our algorithm on the CPU for large datasets which exceed the memory capacity of GPU and use sparse matrix operations which results in large gains in terms of both the memory and the running time requirements. The overall approach is summarized in Algorithm~\ref{alg:otmllvm}.
\begin{algorithm}
\begin{algorithmic}[1]
\REQUIRE Observed data: social network graph or document samples.
\ENSURE Learned latent variable model and infer hidden attributes.
\STATE Estimate the third order moments tensor $M_3$ (implicitly). The tensor is not formed explicitly as we break down the tensor operations into vector and matrix operations.
\STATE Whiten the data, via SVD of $M_2$, to reduce dimensionality via symmetrization and orthogonalization.  The third order moments $M_3$ are whitened as $\mathcal{T}$.
\STATE Use stochastic gradient descent to estimate spectrum of whitened (implicit) tensor $\mathcal{T}$.
\STATE Apply post-processing to obtain the topic-word matrix or the community memberships.
\STATE If ground truth is known, validate the results using various evaluation measures.
\end{algorithmic}
\caption{Overall approach for learning latent variable models via a moment-based approach.}
\label{alg:otmllvm}
\end{algorithm}

\subsection{Dimensionality Reduction and Whitening}\label{sec:DRandWhite}


Whitening step utilizes linear algebraic manipulations to make the tensor symmetric and orthogonal (in expectation).  Moreover, it leads to dimensionality reduction since it (implicitly) reduces tensor $M_3$ of size $O(n^3)$ to a tensor of size $k^3$, where $k$ is the number of communities. Typically we have $k \ll n$. The whitening step also converts the tensor $M_3$ to a symmetric orthogonal tensor. The whitening matrix $W\in \Rbb^{n_A \times k}$ satisfies $W^\top M_2 W = I$. The idea is that if the bilinear projection of the second order moment onto $W$ results in the identity matrix, then a trilinear projection of the third order moment onto $W$ would result in an orthogonal tensor. We  use multilinear operations  to get an orthogonal   tensor $\mathcal{T} :=M_3(W,W,W)$.

The whitening matrix $W$ is computed via truncated $k-$svd of the second order moments.
\begin{equation*}
W = U_{M_2} \Sigma_{M_2}^{-1/2},
\end{equation*}
where $U_{M_2}$ and $\Sigma_{M_2}=\diag(\sigma_{M_2,1},\ldots,\sigma_{M_2,k})$ are the top $k$ singular vectors and singular values of $M_2$ respectively.
We then perform multilinear transformations on the triplet data using the whitening matrix. The whitened data is thus
\begin{align*}
y^t_A  &: = \left<W, {c^t}\right>,\\
 y^t_B &:= \left<W, c^t  \right>,\\
y^t_C &: = \left<W, c^t\right>,
\end{align*} for the topic modeling, where $t$ denotes the index of the documents. Note that $y^t_A$, $y^t_B$ and $y^t_C$ $\in \Rbb^{k}$. Implicitly, the whitened tensor is $\mathcal{T} = \frac{1}{\nx} \sum\limits_{t\in X} y^t_A \otimes y^t_B \otimes y^t_C$ and is a $k\times k \times k$ dimension tensor.
Since  $k \ll n$, the dimensionality reduction is crucial for our speedup.

%

\subsection{Stochastic Tensor Gradient Descent}
\label{sec:sto_ten_grad_des}


In  \citep{AnandkumarEtal:community12COLT} and \citep{AGHKT12}, the power method with deflation is used for tensor decomposition where the eigenvectors are recovered by iterating over multiple loops in a serial manner. Furthermore, batch data is used in their iterative power method which makes that algorithm slower than its stochastic counterpart.  In addition to implementing a stochastic spectral optimization algorithm, we achieve further speed-up by efficiently parallelizing the stochastic updates.  

Let $\mathbf{v}=[v_1|v_2|\ldots|v_k]$ be the true eigenvectors. Denote the cardinality of the sample set as $\nx$, i.e., $\nx:=|X|$.  Now that we have  the whitened tensor, we propose the \emph{Stochastic Tensor Gradient Descent} (STGD) algorithm for  tensor decomposition.
Consider the tensor $\mathcal{T} \in \R^{k \times k \times k}$ using whitened samples, i.e.,
\begin{align*}
\mathcal{T} & = \sum_{t\in X}{\mathcal{T}^t} = \frac{(\alpha_0+1)(\alpha_0+2)}{2\nx}\sum_{t \in X} y^t_A \otimes y^t_B \otimes y^t_C \\
&- \frac{\alpha_0(\alpha_0+1)}{2\nx}\sum_{t\in X} \left[ y^t_A\otimes y^t_B \otimes\bar{y}_C + y^t_A\otimes \bar{y}_B \otimes y^t_C + \bar{y}_A\otimes y^t_B \otimes y^t_C\right] + \alpha_0^2 \bar{y}_A \otimes \bar{y}_B \otimes \bar{y}_C,
\end{align*} where $t\in X$ and denotes the index of the online data and $\bar{y}_A$,  $\bar{y}_B$, and $\bar{y}_C$ denote the mean of the whitened data.
Our goal is to find a symmetric CP decomposition of the whitened tensor.
\begin{definition}
Our optimization problem is given by
\[
 \arg \min_{\mathbf{v}: \| v_i\|_F^2=1} \Big\{ \big\lVert  \sum_{i\in[k]}{ \otimes^3 v_i}- \sum_{t\in X} \mathcal{T}^t\big\lVert_F ^2  + \theta \lVert \sum_{i\in [k]} \otimes^3 v_i \lVert_F^2
 \Big\},
\] where $v_i$ are the unknown components to be estimated, and   $\theta>0$ is some fixed parameter.
\end{definition}
In order to encourage orthogonality between eigenvectors, we have the extra term as $\theta \lVert \sum_{i\in [k]} \otimes^3 v_i \lVert_F^2$. Since $\lVert \sum_{t\in X}{\mathcal{T}^t}\lVert_F^2$ is a constant, the above minimization is the same as minimizing a loss function $L(\mathbf{v}) :=\frac{1}{\nx} \sum_{t} L^t(\mathbf{v})$, where  $L^t(\mathbf{v})$  is the loss function evaluated at node $t\in X$, and  is given by
\begin{equation}\label{eqn:loss1}
L^t(\mathbf{v}): =  \frac{1+\theta}{2} \big\lVert \sum_{i\in [k]} \otimes^3 v_i \big\lVert_F^2 -  \big\langle \sum_{i\in [k]} {\otimes^3 v_i}, {\mathcal{T}^t}\big\rangle
\end{equation} The loss function has two terms, \viz  the term $\lVert \sum_{i\in [k]} \otimes^3 v_i \lVert_F^2$, which can be interpreted as the orthogonality cost,  which we need to minimize,  and the second term  $\langle \sum_{i\in [k]} {\otimes^3 v_i}, {\mathcal{T}^t}\rangle$, which  can be viewed as the correlation reward to be maximized. The parameter $\theta$ provides  additional flexibility for tuning between the two terms.

Let $\Phi^t:=\left[\phi^t_1|\phi^t_2| \ldots | \phi^t_k\right]$ denote the estimation of the eigenvectors using the whitened data point $t$, where $\phi^t_i\in \mathbb{R}^{k},\ i\in[k]$.
Taking the derivative of the loss function leads us to the iterative update equation for the stochastic gradient descent which is
\[
\phi_i^{t+1} \leftarrow \phi_i^t - \beta^t \frac{\partial L^t}{\partial v_i}\llvert_{\phi_i^t}
, \; \forall i \in [k]
\]
 where $\beta^t$ is the learning rate. Computing the derivative of the loss function and substituting the result leads to the following lemma.

\begin{figure}
\centering{
\def\svgwidth{1.4in}
\begingroup%
  \makeatletter%
  \providecommand\color[2][]{%
    \errmessage{(Inkscape) Color is used for the text in Inkscape, but the package 'color.sty' is not loaded}%
    \renewcommand\color[2][]{}%
  }%
  \providecommand\transparent[1]{%
    \errmessage{(Inkscape) Transparency is used (non-zero) for the text in Inkscape, but the package 'transparent.sty' is not loaded}%
    \renewcommand\transparent[1]{}%
  }%
  \providecommand\rotatebox[2]{#2}%
  \ifx\svgwidth\undefined%
    \setlength{\unitlength}{280.73261719bp}%
    \ifx\svgscale\undefined%
      \relax%
    \else%
      \setlength{\unitlength}{\unitlength * \real{\svgscale}}%
    \fi%
  \else%
    \setlength{\unitlength}{\svgwidth}%
  \fi%
  \global\let\svgwidth\undefined%
  \global\let\svgscale\undefined%
  \makeatother%
  \begin{picture}(1,0.71109982)%
    \put(0,0){\includegraphics[width=\unitlength]{\fighome/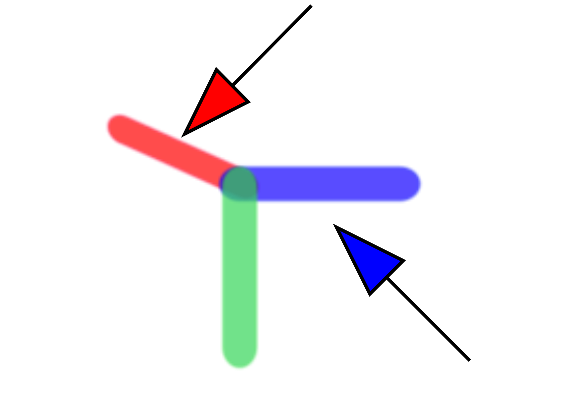}}%
    \put(-0.00283856,0.5550544){\color[rgb]{0,0,0}\makebox(0,0)[lb]{\smash{$y_A^t$}}}%
    \put(0.33059148,0.00806484){\color[rgb]{0,0,0}\makebox(0,0)[lb]{\smash{$y_C^t$}}}%
    \put(0.75804449,0.3785241){\color[rgb]{0,0,0}\makebox(0,0)[lb]{\smash{$y_B^t$}}}%
    \put(0.60359839,0.68511869){\color[rgb]{0,0,0}\makebox(0,0)[lb]{\smash{$v_i^t$}}}%
    \put(0.87041161,0.08212112){\color[rgb]{0,0,0}\makebox(0,0)[lb]{\smash{$v_i^t$}}}%
  \end{picture}%
\endgroup%
}
\caption{Schematic representation of the stochastic updates for the spectral estimation. Note the we never form the tensor explicitly, since the gradient involves vector products by collapsing two modes, as shown in Equation~\ref{eqn:implicit_vec_pdt}.}
\label{fig:stoc_up}
\end{figure}
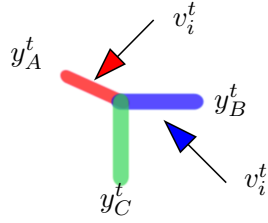

\begin{lemma}
\label{lem:update}
The stochastic updates for the eigenvectors are given by
{\small
\begin{align}
\label{eqn:implicit_vec_pdt}
& \phi_i^{t+1} \leftarrow \phi_i^t  -  \frac{1+\theta}{2} \beta^t \sum\limits_{j=1}^{k} \left[\left<\phi_j^t,\phi_i^t\right>^2 \phi_j^t\right]
 + \beta^t \frac{(\alpha_0+1)(\alpha_0+2)}{2}\left<\phi_i^t, y_A^t\right> \left<\phi_i^t, y_B^t\right> y_C^t +\beta^t \alpha_0^2 \left<\phi_i^t,\bar{y}_A\right>\left<\phi_i^t,\bar{y}_B^t\right>\bar{y}_C \nonumber\\
& -  \beta^t \frac{\alpha_0(\alpha_0+1)}{2}\left<\phi_i^t, y_A^t\right>\left<\phi_i^t, y_B^t\right>\bar{y}_C 
- \beta^t \frac{\alpha_0(\alpha_0+1)}{2}\left<\phi_i^t, y_A^t\right>\left<\phi_i^t, \bar{y}_B\right>y_C 
-  \beta^t \frac{\alpha_0(\alpha_0+1)}{2}\left<\phi_i^t,\bar{y}_A\right>\left<\phi_i^t,y_B^t\right>y_C,
\end{align}
}
\end{lemma}
In Equation~\eqref{eqn:implicit_vec_pdt}, all our tensor operations are in terms of efficient sample vector inner products, and no tensor is explicitly formed. The multilinear operations are shown in Figure~\ref{fig:stoc_up}. We choose $\theta =1$ in our experiments to ensure that there is sufficient penalty for non-orthogonality, which prevents us from obtaining degenerate solutions.

After learning the decomposition of the third order moment, we perform post-processing to estimate $\widehat{\Pi}$.

\subsection{Post-processing}\label{sec:post_process}

Eigenvalues $\Lambda:=[\lambda_1,\lambda_2,\ldots, \lambda_k]$  are estimated as the norm of the eigenvectors $\lambda_i = {\lVert{\phi_i}\rVert}^3$.
\begin{lemma}\label{lemma:postprocessing}
After we obtain $\Lambda$ and $\Phi$, the estimate for the topic-word matrix is given by
\[
\hat{\mu} = {W^\top}^\dag \Phi,
\]
and in the community setting,   the community membership matrix is given by
\[
\hat{\Pi}_{A^c} = 
 \diag (\gamma)^{1/3}\diag(\Lambda)^{-1} \Phi^\top\hat{
W}^\top G_{A,A^c}.
\]
 where $A^c : = X \cup B \cup C$. Similarly, we estimate $\hat{\Pi}_A$ by exchanging the roles of $X$ and $A$. Next, we obtain the Dirichlet distribution parameters
\begin{equation*}
\hat{\alpha_i} =  \gamma^2\lambda_i^{-2},  
\forall i \in [k].
\end{equation*}\end{lemma}
 where $\gamma^2$ is chosen such that we have normalization$
\sum_{i\in[k]}\hat{ \alpha}_i : =\sum_{i\in [k]}\frac{\alpha_i}{\alpha_0}=1.
$

Thus, we perform STGD method to estimate the eigenvectors and eigenvalues of the whitened tensor, and then use these to estimate the topic word matrix $\mu$ and community membership matrix $\widehat{\Pi}$ by thresholding.

\section{Implementation Details}

\subsection{Symmetrization Step to Compute $M_2$}
\begin{figure*}[h]
\centering
{\begin{minipage}{4in}
\centering
\def\svgwidth{\textwidth}
\begingroup%
  \makeatletter%
  \providecommand\color[2][]{%
    \errmessage{(Inkscape) Color is used for the text in Inkscape, but the package 'color.sty' is not loaded}%
    \renewcommand\color[2][]{}%
  }%
  \providecommand\transparent[1]{%
    \errmessage{(Inkscape) Transparency is used (non-zero) for the text in Inkscape, but the package 'transparent.sty' is not loaded}%
    \renewcommand\transparent[1]{}%
  }%
  \providecommand\rotatebox[2]{#2}%
  \ifx\svgwidth\undefined%
    \setlength{\unitlength}{747.20448486bp}%
    \ifx\svgscale\undefined%
      \relax%
    \else%
      \setlength{\unitlength}{\unitlength * \real{\svgscale}}%
    \fi%
  \else%
    \setlength{\unitlength}{\svgwidth}%
  \fi%
  \global\let\svgwidth\undefined%
  \global\let\svgscale\undefined%
  \makeatother%
  \begin{picture}(1,0.23765382)%
    \put(0,0){\includegraphics[width=\unitlength]{\fighome/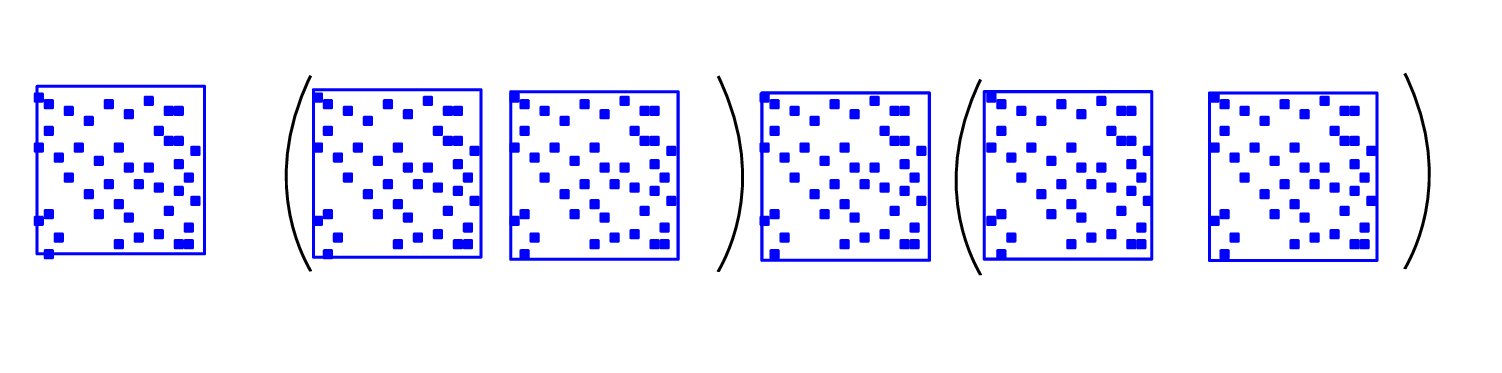}}%
    \put(0.15651648,0.12646764){\color[rgb]{0,0,0}\makebox(0,0)[lb]{\smash{$=$}}}%
    \put(0.46220125,0.18491793){\color[rgb]{0,0,0}\makebox(0,0)[lb]{\smash{$\dag$}}}%
    \put(0.9106669,0.18384786){\color[rgb]{0,0,0}\makebox(0,0)[lb]{\smash{$\top$}}}%
    \put(0.76865456,0.18491793){\color[rgb]{0,0,0}\makebox(0,0)[lb]{\smash{$\dag$}}}%
    \put(0.78514754,0.18384786){\color[rgb]{0,0,0}\makebox(0,0)[lb]{\smash{$\top$}}}%
    \put(-0.01606648,0.13703326){\color[rgb]{0,0,0}\makebox(0,0)[lb]{\smash{$\lvert A\rvert$}}}%
    \put(0.06959692,0.19913139){\color[rgb]{0,0,0}\makebox(0,0)[lb]{\smash{$\lvert A\rvert$}}}%
  \end{picture}%
\endgroup%
\end{minipage}}
\hfil
{\begin{minipage}{4in}
\centering
\def\svgwidth{\textwidth}
\begingroup%
  \makeatletter%
  \providecommand\color[2][]{%
    \errmessage{(Inkscape) Color is used for the text in Inkscape, but the package 'color.sty' is not loaded}%
    \renewcommand\color[2][]{}%
  }%
  \providecommand\transparent[1]{%
    \errmessage{(Inkscape) Transparency is used (non-zero) for the text in Inkscape, but the package 'transparent.sty' is not loaded}%
    \renewcommand\transparent[1]{}%
  }%
  \providecommand\rotatebox[2]{#2}%
  \ifx\svgwidth\undefined%
    \setlength{\unitlength}{720.00027734bp}%
    \ifx\svgscale\undefined%
      \relax%
    \else%
      \setlength{\unitlength}{\unitlength * \real{\svgscale}}%
    \fi%
  \else%
    \setlength{\unitlength}{\svgwidth}%
  \fi%
  \global\let\svgwidth\undefined%
  \global\let\svgscale\undefined%
  \makeatother%
  \begin{picture}(1,0.24667124)%
    \put(0,0){\includegraphics[width=\unitlength]{\fighome/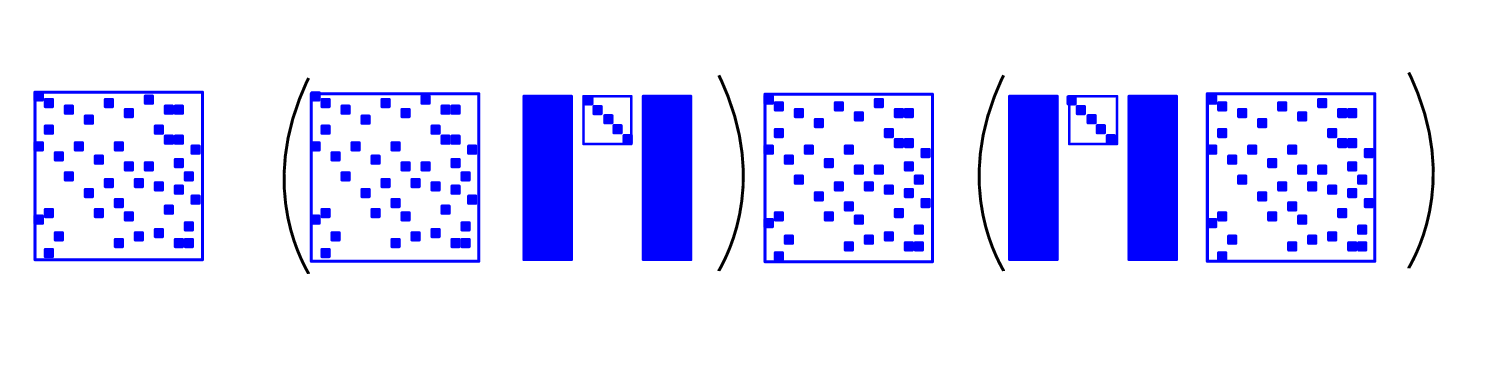}}%
    \put(0.14428852,0.12068267){\color[rgb]{0,0,0}\makebox(0,0)[lb]{\smash{$=$}}}%
    \put(0.45436066,0.1765092){\color[rgb]{0,0,0}\makebox(0,0)[lb]{\smash{$\top$}}}%
    \put(0.77777514,0.18445862){\color[rgb]{0,0,0}\makebox(0,0)[lb]{\smash{$\top$}}}%
    \put(0.91555287,0.18445862){\color[rgb]{0,0,0}\makebox(0,0)[lb]{\smash{$\top$}}}%
  \end{picture}%
\endgroup%
\end{minipage}}
\hfil
{\begin{minipage}{4in}
\centering
\def\svgwidth{\textwidth}
\begingroup%
  \makeatletter%
  \providecommand\color[2][]{%
    \errmessage{(Inkscape) Color is used for the text in Inkscape, but the package 'color.sty' is not loaded}%
    \renewcommand\color[2][]{}%
  }%
  \providecommand\transparent[1]{%
    \errmessage{(Inkscape) Transparency is used (non-zero) for the text in Inkscape, but the package 'transparent.sty' is not loaded}%
    \renewcommand\transparent[1]{}%
  }%
  \providecommand\rotatebox[2]{#2}%
  \ifx\svgwidth\undefined%
    \setlength{\unitlength}{720.00075303bp}%
    \ifx\svgscale\undefined%
      \relax%
    \else%
      \setlength{\unitlength}{\unitlength * \real{\svgscale}}%
    \fi%
  \else%
    \setlength{\unitlength}{\svgwidth}%
  \fi%
  \global\let\svgwidth\undefined%
  \global\let\svgscale\undefined%
  \makeatother%
  \begin{picture}(1,0.24663905)%
    \put(0,0){\includegraphics[width=\unitlength]{\fighome/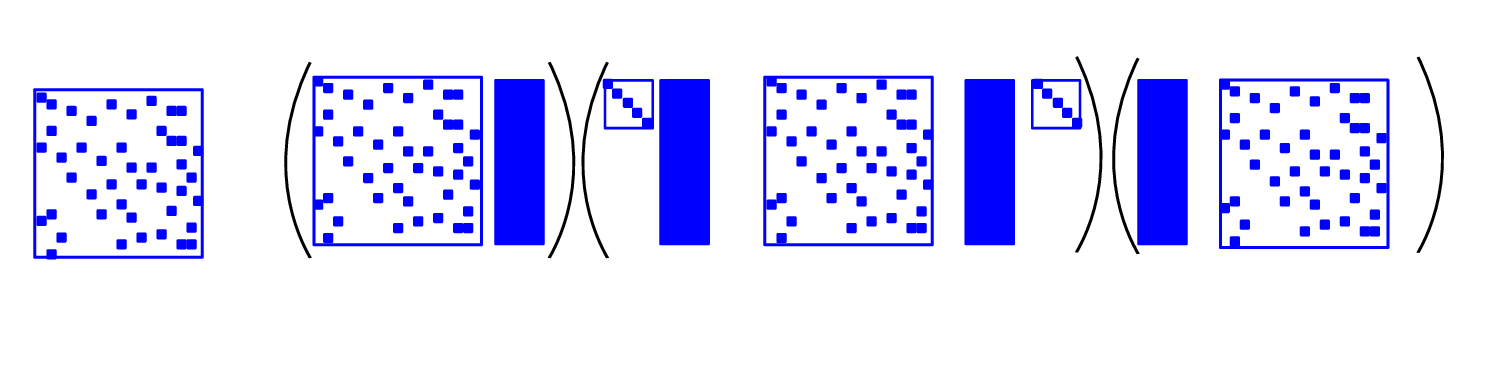}}%
    \put(0.14411109,0.13121129){\color[rgb]{0,0,0}\makebox(0,0)[lb]{\smash{$=$}}}%
    \put(0.47093097,0.19135582){\color[rgb]{0,0,0}\makebox(0,0)[lb]{\smash{$\top$}}}%
    \put(0.78426398,0.19580026){\color[rgb]{0,0,0}\makebox(0,0)[lb]{\smash{$\top$}}}%
    \put(0.91981939,0.19357804){\color[rgb]{0,0,0}\makebox(0,0)[lb]{\smash{$\top$}}}%
  \end{picture}%
\endgroup%
\end{minipage}}
\caption{By performing the matrix multiplications in an efficient order (Equation~\eqref{eq:2moment_graph}), we avoid products involving $O(n) \times O(n)$ objects. Instead, we use objects of size $O(n) \times k$ which improves the speed, since $k \ll n$. Equation~\eqref{eq:2moment_graph} is equivalent to $ M_2 = $ $\left(\Pairs_{A,B} \Pairs_{C,B}^\dag\right)$ $\Pairs_{C,B}$ $\left(\Pairs_{B,C}^\dag\right)^\top \Pairs_{A,C}^\top$  $ -\text{shift}$, where the $\text{shift}=\frac{\alpha_0}{\alpha_0+1} \left({M_1  }{M_1  }^\top- \diag\left({M_1  } {M_1  }^\top\right) \right)$. We do not explicitly calculate the pseudoinverse but maintain the low rank matrix decomposition form.}\label{fig:dim_reduc}
\end{figure*}

Note that for the topic model, the second order moment $M_2$ can be computed easily from the word-frequency vector. On the other hand, for the community setting, computing $M_2$ requires additional linear algebraic operations. It requires computation of matrices $Z_B$ and $Z_C$ in equation~\eqref{eq:transitionMat}. This requires computation of pseudo-inverses of ``Pairs'' matrices.
Now, note that pseudo-inverse of $\left(\Pairs\left(B,C\right)\right)$ in Equation~\eqref{eq:transitionMat} can be computed using rank $k$-SVD:
\begin{align*}
& \text{k-SVD}\left(\Pairs\left(B,C\right)\right) = U_B(:,1:k) \Sigma_{BC}(1:k) V_C(:,1:k)^\top.
\end{align*}We exploit the low rank property to have efficient running times and storage. We first implement the k-SVD of Pairs, given by $G_{X,C}^\top G_{X,B} $. Then the order in which the matrix products are carried out plays a significant role in terms of both memory and speed. Note that  $Z_C$ involves the multiplication of a sequence of matrices of sizes $\mathbb{R}^{n_A\times n_B}$, $\mathbb{R}^{n_B\times k}$, $\mathbb{R}^{k\times k}$, $\mathbb{R}^{k\times n_C}$,  $G_{x,C}^\top G_{x,B} $ involves products of sizes $\mathbb{R}^{n_C\times k}$,  $\mathbb{R}^{k\times k}$, $\mathbb{R}^{k\times n_B}$, and $Z_B$ involving products of sizes $\mathbb{R}^{n_A\times n_C}$, $\mathbb{R}^{n_C\times k}$, $\mathbb{R}^{k\times k}$, $\mathbb{R}^{k\times n_B}$. While performing these products, we avoid products of sizes $\mathbb{R}^{O(n)\times O(n)}$ and $\mathbb{R}^{O(n)\times O(n)}$. This allows us to have efficient storage requirements. Such manipulations are represented in Figure~\ref{fig:dim_reduc}.

We then orthogonalize the third order moments to reduce the dimension of its modes to $k$.
We perform linear transformations on the data corresponding to the partitions $A$, $B$ and $C$ using the whitening matrix.
The whitened data is thus $y^t_A  : = \left<W, G^\top_{t,A}\right>$, $y^t_B := \left<W, Z_B G^\top_{t,B}\right>$, and
$y^t_C : = \left<W, Z_C G^\top_{t,C}\right>$, where $t\in X$ and denotes the index of the online data.
Since  $k \ll n$, the dimensionality reduction is crucial for our speedup.

\subsection{Efficient Randomized SVD Computations}\label{sec:Apdx_sparse}
When we consider very large-scale data, the whitening matrix is a bottleneck to handle when we aim for fast running times. We obtain the low rank approximation of matrices using random projections. In the CPU implementation, we use \emph{tall-thin SVD} (on a sparse matrix) via the Lanczos algorithm after the projection and in the GPU implementation, we use \emph{tall-thin QR}. 
We give the overview of these methods below. Again, we use graph community membership model without loss of generality.

\paragraph{Randomized low rank approximation: }
From~\citep{gittens2013revisiting}, for the $k$-rank positive semi-definite matrix ${M_2} \in \mathbb{R}^{n_A \times n_A}$ with $n_A \gg k$, we can perform random projection to reduce dimensionality. More precisely, if we have a random matrix $S\in \mathbb{R}^{n_A \times \tilde{k}}$ with unit norm (rotation matrix), we project $M_2 $ onto this random matrix to get $\mathbb{R}^{n\times \tilde{k}}$ tall-thin matrix. Note that we choose $\tilde{k}=2k$ in our implementation.  We will obtain lower dimension approximation of $M_2$ in $\mathbb{R}^{\tilde{k} \times \tilde{k}}$.
Here we emphasize that $S\in \mathbb{R}^{n\times \tilde{k}}$ is a random matrix for dense ${M_2} $. However for sparse ${M_2} $, $S\in \{0,1\}^{n\times \tilde{k}}$ is a column selection matrix with random sign for each entry.

After the projection, one approach we use is SVD on this tall-thin ($\mathbb{R}^{n\times \tilde{k}}$) matrix. Define $O:= {M_2}  S \in \mathbb{R}^{n \times \tilde{k}}$ and $\Omega := S^\top {M_2} S \in \mathbb{R}^{\tilde{k} \times \tilde{k}}$. A low rank approximation of ${M_2} $ is given by $O\Omega^\dag O^\top$~\citep{gittens2013revisiting}.
Recall that the definition of a whitening matrix $W$ is that $W^\top {M_2}  W = I$. We can obtain the whitening matrix of ${M_2} $ without directly doing a SVD on ${M_2}  \in \mathbb{R}^{n_A \times n_A}$.

\smallskip
\emph{Tall-thin SVD: }This is used in the CPU implementation.
The whitening matrix can be obtained by
\begin{equation}\label{eq:nystrom_whiten}
W \approx(O^\dag)^\top (\Omega^{\frac{1}{2}})^\top.
\end{equation}
The pseudo code for computing the whitening matrix $W$ using tall-thin SVD is given in Algorithm~\ref{alg:pinv}.
\begin{algorithm}
\caption{Randomized Tall-thin SVD}
\label{alg:pinv}
\begin{algorithmic}[1]
\REQUIRE Second moment matrix $M_2$.
\ENSURE Whitening matrix $W$.
\STATE Generate random matrix $S\in \mathbb{R}^{n\times \tilde{k}}$ if  $M_2 $ is dense.
\STATE Generate column selection matrix with random sign $S\in \{0,1\}^{n\times \tilde{k}}$ if $M_2 $ is sparse.
\STATE $O = M_2 S\in \mathbb{R}^{n \times \tilde{k}}$
\STATE $[U_O, L_O, V_O] =$SVD$(O)$
\STATE $\Omega = S^\top O\in \mathbb{R}^{\tilde{k} \times \tilde{k}}$
\STATE $[U_\Omega, L_\Omega, V_\Omega] = $SVD$(\Omega)$
\STATE $W = U_O L_O^{-1} V_O^\top V_\Omega L_\Omega^{\frac{1}{2}} U_\Omega^\top$
\end{algorithmic}
\end{algorithm}
Therefore, we only need to compute SVD of a tall-thin matrix $O\in\mathbb{R}^{n_A \times \tilde{k}}$.  Note that $\Omega \in \mathbb{R}^{\tilde{k} \times \tilde{k}}$, its square-root is easy to compute.
Similarly, pseudoinverses can also be obtained without directly doing SVD. For instance, the pseudoinverse of the $\Pairs\left(B,C\right)$ matrix is given by
\[
\left(\Pairs\left(B,C\right)\right)^\dag = (J^\dag)^\top \Psi J^\dag,
\]
where $\Psi = S^\top \left(\Pairs\left(B,C\right)\right) S$ and $J = \left(\Pairs\left(B,C\right)\right) S$.
The pseudo code for computing pseudoinverses is given in Algorithm~\ref{alg:ttsvd}.
\begin{algorithm}
\caption{Randomized Pseudoinverse}
\label{alg:ttsvd}
\begin{algorithmic}[1]
\REQUIRE Pairs matrix $\Pairs\left(B,C\right)$.
\ENSURE Pseudoinverse of the pairs matrix $\left(\Pairs\left(B,C\right)\right)^\dag$.
\STATE Generate random matrix $S\in \mathbb{R}^{n,k}$ if  $M_2 $ is dense.
\STATE Generate column selection matrix with random sign $S\in \{0,1\}^{n\times k}$ if $M_2 $ is sparse.
\STATE $J = \left(\Pairs\left(B,C\right)\right) S$
\STATE $\Psi = S^\top J$
\STATE $[U_J, L_J, V_J] =$SVD$(J)$
\STATE $\left(\Pairs\left(B,C\right)\right)^\dag = U_J L_J^{-1}V_J^\top \Psi V_J L_J^{-1}U_J^\top$
\end{algorithmic}
\end{algorithm}

The sparse representation of the data allows for scalability on a single machine to datasets having millions of nodes.
Although the GPU has SIMD architecture which makes parallelization efficient, it lacks advanced libraries with sparse SVD operations and out-of-GPU-core implementations. We therefore implement the sparse format on CPU for sparse datasets. We implement our algorithm using random projection for efficient dimensionality reduction~\citep{DBLP:journals/corr/abs-1207-6365} along with the sparse matrix operations available in the Eigen toolkit\footnote{\scriptsize{\url{http://eigen.tuxfamily.org/index.php?title=Main_Page}}}, and we use the SVDLIBC~\citep{svdlibc2002} library to compute sparse SVD via the Lanczos algorithm.
Theoretically, the Lanczos algorithm~\citep{zbMATH06159604} on a $n \times n$ matrix takes around $(2d+8)n$ flops for a single step where $d$ is the average number of non-zero entries per row.


\smallskip
\emph{Tall-thin QR: }
This is used in the GPU implementation due to the lack of library to do sparse tall-thin SVD. The difference is that we instead implement a tall-thin QR on $O$,  therefore the whitening matrix is obtained as
\[
W \approx Q (R^\dag)^\top (\Omega^{\frac{1}{2}})^\top.
 \]

The main bottleneck for our GPU implementation is device storage, since GPU memory is highly limited and not expandable. Random projections help in reducing the dimensionality from $O(n \times n)$ to $O(n \times k)$ and hence, this fits the data in the GPU memory better. Consequently, after the whitening step, we project the data into $k$-dimensional space. Therefore, the STGD step is dependent only on $k$, and hence can be fit in the GPU memory. So, the main bottleneck is computation of large SVDs. In order to support larger datasets such as the DBLP dataset 
which exceed the GPU memory capacity, we extend our implementation with out-of-GPU-core matrix operations and the Nystrom method~\citep{gittens2013revisiting} for the whitening matrix computation and the pseudoinverse computation in the pre-processing module. 



\subsection{Stochastic updates}\label{sec:stgd}

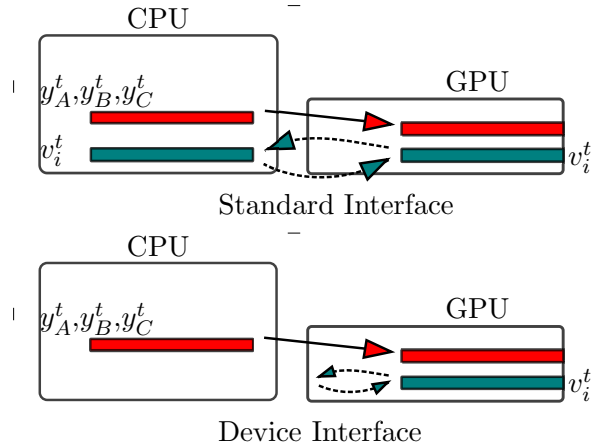
\begin{figure}
\centering{
\def\svgwidth{3.2in}
\begingroup%
  \makeatletter%
  \providecommand\color[2][]{%
    \errmessage{(Inkscape) Color is used for the text in Inkscape, but the package 'color.sty' is not loaded}%
    \renewcommand\color[2][]{}%
  }%
  \providecommand\transparent[1]{%
    \errmessage{(Inkscape) Transparency is used (non-zero) for the text in Inkscape, but the package 'transparent.sty' is not loaded}%
    \renewcommand\transparent[1]{}%
  }%
  \providecommand\rotatebox[2]{#2}%
  \ifx\svgwidth\undefined%
    \setlength{\unitlength}{784.8bp}%
    \ifx\svgscale\undefined%
      \relax%
    \else%
      \setlength{\unitlength}{\unitlength * \real{\svgscale}}%
    \fi%
  \else%
    \setlength{\unitlength}{\svgwidth}%
  \fi%
  \global\let\svgwidth\undefined%
  \global\let\svgscale\undefined%
  \makeatother%
  \begin{picture}(1,0.36799185)%
    \put(0,0){\includegraphics[width=\unitlength]{\fighome/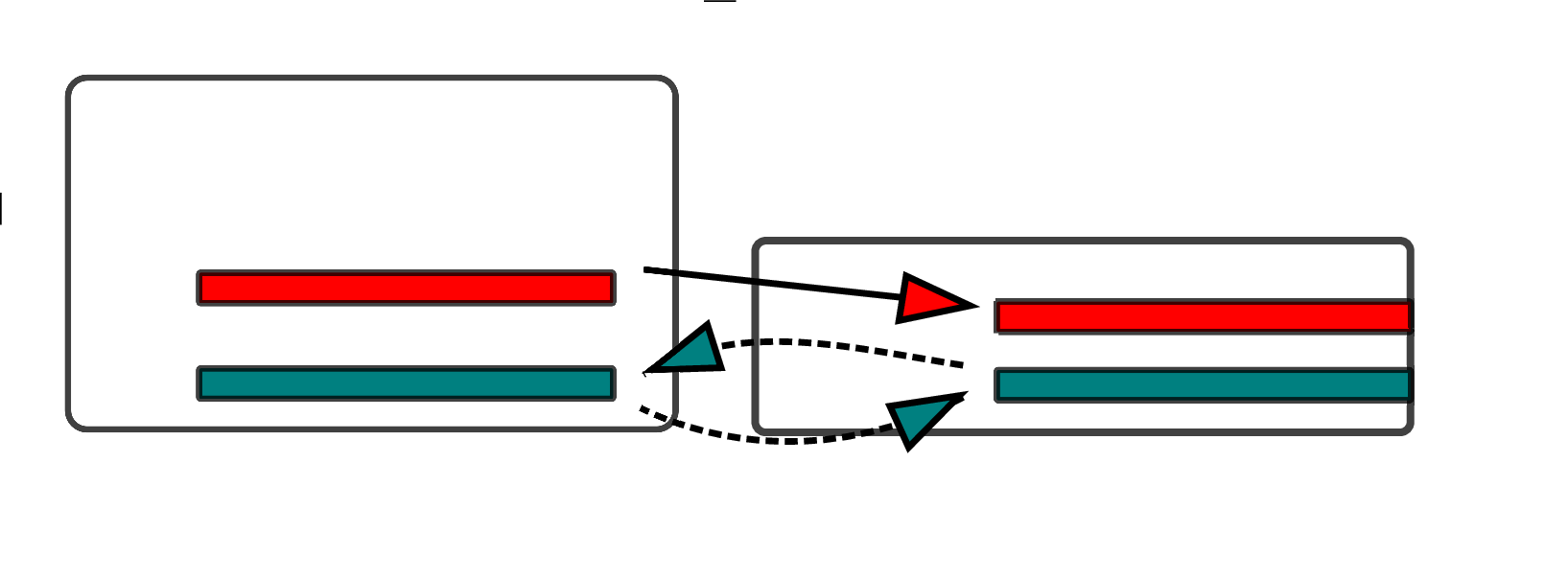}}%
    \put(0.04335005,0.11879389){\color[rgb]{0,0,0}\makebox(0,0)[lb]{\smash{$v_i^t$}}}%
    \put(0.04314793,0.21615305){\color[rgb]{0,0,0}\makebox(0,0)[lb]{\smash{$y_A^t$,$y_B^t$,$y_C^t$}}}%
    \put(0.18612624,0.33110272){\color[rgb]{0,0,0}\makebox(0,0)[lb]{\smash{CPU}}}%
    \put(0.7069755,0.22916592){\color[rgb]{0,0,0}\makebox(0,0)[lb]{\smash{GPU}}}%
    \put(0.33555723,0.02770536){\color[rgb]{0,0,0}\makebox(0,0)[lb]{\smash{Standard Interface}}}%
    \put(0.90796023,0.10734348){\color[rgb]{0,0,0}\makebox(0,0)[lb]{\smash{$v_i^t$}}}%
  \end{picture}%
\endgroup%
}

\centering{
\def\svgwidth{3.2in}
\begingroup%
  \makeatletter%
  \providecommand\color[2][]{%
    \errmessage{(Inkscape) Color is used for the text in Inkscape, but the package 'color.sty' is not loaded}%
    \renewcommand\color[2][]{}%
  }%
  \providecommand\transparent[1]{%
    \errmessage{(Inkscape) Transparency is used (non-zero) for the text in Inkscape, but the package 'transparent.sty' is not loaded}%
    \renewcommand\transparent[1]{}%
  }%
  \providecommand\rotatebox[2]{#2}%
  \ifx\svgwidth\undefined%
    \setlength{\unitlength}{784.8bp}%
    \ifx\svgscale\undefined%
      \relax%
    \else%
      \setlength{\unitlength}{\unitlength * \real{\svgscale}}%
    \fi%
  \else%
    \setlength{\unitlength}{\svgwidth}%
  \fi%
  \global\let\svgwidth\undefined%
  \global\let\svgscale\undefined%
  \makeatother%
  \begin{picture}(1,0.36799185)%
    \put(0,0){\includegraphics[width=\unitlength]{\fighome/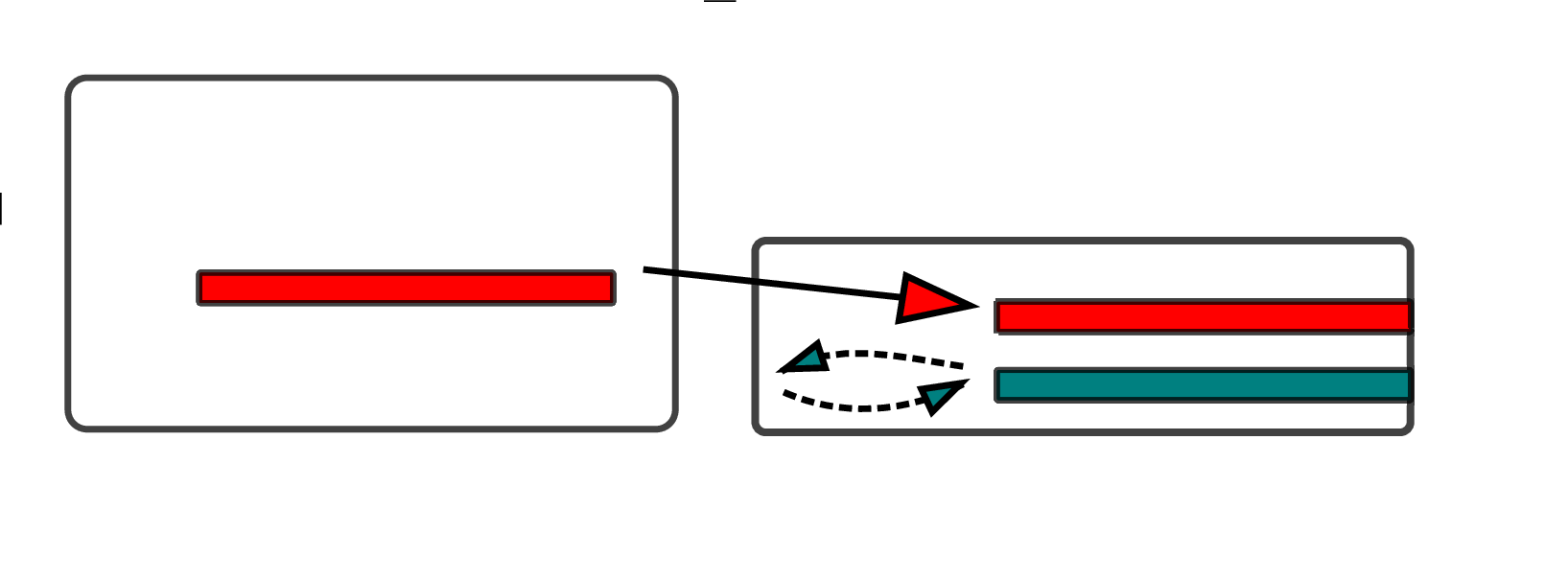}}%
    \put(0.04314793,0.21615305){\color[rgb]{0,0,0}\makebox(0,0)[lb]{\smash{$y_A^t$,$y_B^t$,$y_C^t$}}}%
    \put(0.18612624,0.33110272){\color[rgb]{0,0,0}\makebox(0,0)[lb]{\smash{CPU}}}%
    \put(0.7069755,0.22916592){\color[rgb]{0,0,0}\makebox(0,0)[lb]{\smash{GPU}}}%
    \put(0.33490021,0.02770536){\color[rgb]{0,0,0}\makebox(0,0)[lb]{\smash{Device Interface}}}%
    \put(0.90796023,0.10734348){\color[rgb]{0,0,0}\makebox(0,0)[lb]{\smash{$v_i^t$}}}%
  \end{picture}%
\endgroup%
}
\caption{Data transfers in the standard and device interfaces of the GPU implementation.}
\label{fig:standard_device}
\end{figure}

STGD can potentially be the most computationally intensive task if carried out naively since the storage and manipulation of a $O(n^3)$-sized tensor makes the method not scalable. However we   overcome this problem since we never form the tensor explicitly; instead, we collapse the tensor modes implicitly as shown in Figure~\ref{fig:stoc_up}. We gain large speed up by optimizing the implementation of STGD.
To implement the tensor operations efficiently we convert them into matrix and vector operations so that they are implemented using BLAS routines.  We obtain whitened   vectors $y_A, y_B$ and $y_C$ and manipulate these vectors efficiently to obtain tensor eigenvector updates  using the gradient scaled by a suitable learning rate.

\paragraph{Efficient STGD via stacked vector operations: }
We convert the BLAS II into BLAS III operations by stacking the vectors to form matrices, leading to more efficient operations.
Although the updating equation for the stochastic gradient update is presented serially in Equation~\eqref{eqn:implicit_vec_pdt}, we can update the $k$ eigenvectors simultaneously in parallel. The basic idea is to stack the $k$ eigenvectors $\phi_i\in\mathbb{R}^k$ into a matrix $\mathbf{\Phi}$, then using the internal parallelism designed for BLAS III operations.

Overall, the STGD step involves $1+k+i(2+3k)$ BLAS II over $\mathbb{R}^k$ vectors, 7N BLAS III over $\mathbb{R}^{k\times k}$ matrices and 2 QR operations over $\mathbb{R}^{k \times k}$ matrices, where $i$ denotes the number of iterations. We  provide a count of BLAS operations for various steps in Table~\ref{tab:blas_count}.

\begin{table}[htbp]
 \centering
   \begin{tabular}{@{} |l|c|c|c|c|c| @{}}
\hline
Module & BLAS I & BLAS II & BLAS III & SVD & QR\\
\hline
\hline
Pre & $0$ & $8$ & $19$ & $3$ & $0$\\
STGD & 0 & $Nk$ & $7N$ & $0$ & $2$\\
Post & $0$ & $0$ & $7$ & $0$ & $0$\\
\hline
   \end{tabular}
   \caption{Linear algebraic operation counts: $N$ denotes the number of iterations for STGD and $k$, the number of communities. 
   }
   \label{tab:blas_count}
\end{table}

\paragraph{Reducing communication in GPU implementation: }
In STGD, note that the storage needed for the iterative part does not depend on the number of nodes in the dataset, rather, it depends on the parameter $k$, i.e., the number of communities to be estimated, since whitening performed before STGD leads to dimensionality reduction.  This makes it suitable for storing the required buffers in the GPU memory, and using the CULA device interface for the BLAS operations. In Figure~\ref{fig:standard_device}, we illustrate the data transfer involved in the GPU standard and device interface codes. While the standard interface involves data transfer (including whitened neighborhood vectors and the eigenvectors) at each stochastic iteration between the CPU memory and the GPU memory, the device interface involves allocating and retaining the eigenvectors at each stochastic iteration which in turn speeds up the spectral estimation.

\begin{figure}
\centering
\psfrag{Number of communities}[l]{{Number of communities $k$}}
\psfrag{Time (in seconds) for 100 stochastic iterations}[l]{{ Running time(secs)}}
\psfrag{Scaling of the stochastic algorithm with the rank of the tensor}[c]{}
\psfrag{MATLAB Tensor Toolbox}[l]{{\textbf{MATLAB Tensor Toolbox}}}
\psfrag{CULA Standard Interface}[l]{{\textbf{CULA Standard Interface}}}
\psfrag{CULA Device Interface}[l]{{\textbf{CULA Device Interface}}}
\psfrag{Eigen Sparse}[l]{{\textbf{Eigen Sparse}}}
\includegraphics[width=4.5in]{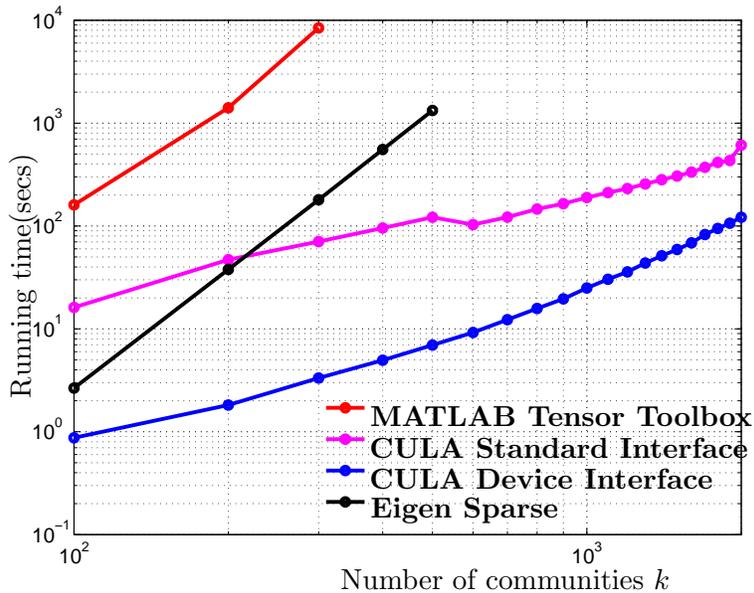}
\caption{Comparison of the running time for STGD under different $k$ for $100$ iterations.}
\label{fig:k_vs_t}
\end{figure}

We compare the running time of  the CULA device code  with the MATLAB code (using the tensor toolbox~\citep{TTB_Software}), CULA standard code and Eigen sparse code in Figure~\ref{fig:k_vs_t}. 
As expected, the GPU implementations of matrix operations are much faster and scale much better than the CPU implementations. Among the CPU codes, we notice that sparsity and optimization offered by the Eigen toolkit gives us huge gains.
We obtain orders of magnitude of speed up for the GPU device code as we place the buffers in the GPU memory and transfer minimal amount of data involving the whitened vectors only once at the beginning of each iteration.  
The running time for the CULA standard code is more than the device code because of the CPU-GPU data transfer overhead. For the same reason, the sparse CPU implementation, by avoiding the data transfer overhead, performs better than the GPU standard code for very small number of communities. We note that there is no performance degradation due to the parallelization of the matrix operations. After whitening, the STGD requires the most code design and optimization effort, and so we convert that into BLAS-like routines.

\subsection{Computational Complexity}

\begin{table}[htbp]
 \centering
   \begin{tabular}{@{} |l|l|l| @{}}
\hline
Module & Time & Space\\
\hline
\hline
Pre-processing (Matrix Multiplication) & $O\left(\max(nsk/c, \log s)\right)$ & $O\left(\max(s^2,sk)\right)$\\
\hline
Pre-processing (CPU SVD) & $O\left(\max(nsk/c,\log s) + \max(k^2/c,k) \right)$ & $O(sk)$\\
\hline
Pre-processing (GPU QR) & $O\left( \max(sk^2/c, \log s) + \max(sk^2/c,\log k) \right)$ & $O(sk)$\\
\hline
Pre-processing(short-thin SVD) & $O\left(\max(k^3/c,\log k) +\max(k^2/c,k)\right)$ & $O(k^2)$\\
\hline
STGD & $O\left(\max(k^3/c, \log k)   \right)$ & $O(k^2)$\\
\hline
Post-processing & $O\left(\max(nsk/c, \log s)\right)$ & $O(nk)$\\
\hline
   \end{tabular}
   \caption{The time and space complexity (number of compute cores required) of our algorithm. Note that $k \ll n$, $s$ is the average degree of a node (or equivalently, the average number of non-zeros per row/column in the adjacency sub-matrix); note that the STGD time is per iteration time. We denote the number of cores as $c$ - the time-space trade-off depends on this parameter.}
   \label{tab:complexity}
\end{table}

We partition the execution of our algorithm into three main modules namely, pre-processing, STGD and post-processing, whose various matrix operation counts  are listed above in Table~\ref{tab:blas_count}.

The theoretical asymptotic complexity of our method is summarized in Table~\ref{tab:complexity} and is best addressed by considering the parallel model of computation~\citep{jaja1992introduction}, i.e., wherein a number of processors or compute cores are operating on the data simultaneously in parallel. This is justified considering that we implement our method on GPUs and matrix products are embarrassingly parallel. Note that this is different from serial computational complexity. We now break down the entries in Table~\ref{tab:complexity}. First, we recall a basic lemma regarding the lower bound on the time complexity for parallel addition along with the required number of cores to achieve a speed-up.
\begin{lemma}
~\citep{jaja1992introduction}
\label{lem:add}
Addition of $s$ numbers in serial takes $O(s)$ time; with $\Omega(s / \log s )$ cores, this can be improved to $O( \log s)$ time in the best case.
\end{lemma}
Essentially, this speed-up is achieved by recursively adding pairs of numbers in parallel. 
\begin{lemma}
~\citep{jaja1992introduction}
\label{lem:mat_mul}
Consider $M \in \mathbb{R}^{p \times q}$ and $N \in \mathbb{R}^{q \times r}$ with $s$ non-zeros per row/column. Naive serial matrix multiplication requires $O(psr)$ time; with $\Omega(psr / \log s)$ cores, this can be improved to $O( \log s)$ time in the best case.
\end{lemma}
Lemma~\ref{lem:mat_mul} follows by simply parallelizing the sparse inner products and applying Lemma~\ref{lem:add} for the addition in the inner products. Note that, this can be generalized to the fact that given $c$ cores, the multiplication can be performed in $O(\max(psr/c, \log s))$ running time.
\subsubsection{Pre-processing}

\paragraph{Random projection: }In preprocessing, given $c$ compute cores, we first do random projection using matrix multiplication. We multiply an $O(n) \times O(n)$ matrix $M_2$ with  an $O(n) \times O(k)$ random matrix $S$. Therefore, this requires $O(nsk)$ serial operations, where $s$ is the number of non-zero elements per row/column of $M_2$.  Using Lemma~\ref{lem:mat_mul}, given $c = \frac{nsk}{\log s}$ cores, we could achieve $O(\log s)$ computational complexity. However, the parallel computational complexity is not further reduced with more than $\frac{nsk}{\log s}$ cores.

After the multiplication, we use \emph{tall-thin SVD} for CPU implementation, and \emph{tall-thin QR} for GPU implementation.

\paragraph{Tall-thin SVD: }We perform Lanczos SVD on the tall-thin sparse $O(n)\times O(k)$ matrix, which involves a tri-diagonalization followed with the QR on the tri-diagonal matrix. Given $c = \frac{nsk}{\log s}$ cores, the computational complexity of the tri-diagonalization is $O(\log s)$. We then do QR on the tridiagonal matrix which is as cheap as $O(k^2)$ serially. Each orthogolization requires $O(k)$ inner products of constant entry vectors, and there are $O(k)$ such orthogolizations to be done. Therefore given $O(k)$ cores, the complexity is               $O(k)$.  More cores does not help since the degree of parallelism is $k$.

\paragraph{Tall-thin QR: }Alternatively, we perform QR in the GPU implementation which takes $O(sk^2)$. To arrive at the complexity of obtaining $Q$, we analyze the Gram-Schmidt orthonormalization procedure under sparsity and parallelism conditions. Consider a serial Gram-Schmidt on $k$ columns (which are $s$-dense) of $O(n) \times O(k)$ matrix. For each of the columns $2$ to $k$, we perform projection on the previously computed components and subtract it. Both inner product and subtraction operations are on the $s$-dense columns and there are $O(s)$ operations which are done $O(k^2)$ times serially. The last step is the normalization of $k$ $s$-dense vectors with is an $O(sk)$ operation. This leads to a serial complexity of $O(sk^2 + sk) = O(sk^2)$. Using this, we may obtain the parallel complexity in different regimes of the number of cores as follows.

\emph{Parallelism for inner products }: For each component $i$, we need $i-1$ projections on previous components which can be parallel.  Each projection involves scaling and inner product operations on a pair of $s$-dense vectors.  Using Lemma~\ref{lem:add},  projection for component $i$ can be performed in $O(\max(\frac{sk}{c}, \log s) )$ time.  $O(\log s)$ complexity is obtained using $O(sk/\log s)$ cores.

\emph{Parallelism for subtractions}:  For each component $i$, we need $i-1$ subtractions on a $s$-dense vector after the projection. Serially the subtraction requires $O(sk)$ operations, and this can be reduced to $O(\log k)$ with $O(sk/\log k)$ cores in the best case. The complexity is $O(\max(\frac{sk}{c}, \log k) )$.

Combing the inner products and subtractions, the complexity is $O\left(\max(\frac{sk}{c}, \log s) +\max(\frac{sk}{c}, \log k)\right)$ for component $i$. There are $k$ components in total, which can not be parallel. In total, the complexity for the parallel QR is $O\left(\max(\frac{sk^2}{c}, \log s) +\max(\frac{sk^2}{c}, \log k)\right)$.

\paragraph{Short-thin SVD: }SVD of the smaller $O( \Rbb^{k \times k})$ matrix  time requires $O(k^3)$ computations in serially. We note that this is the bottleneck for the computational complexity, but we emphasize that $k$ is sufficiently small in many applications. Furthermore, this $k^3$ complexity  can be reduced by using distributed SVD algorithms~e.g.~\citep{kannan2014principal,feldman2013turning}. An analysis with respect to Lanczos parallel SVD is similar with the discussion in the Tall-thin SVD paragraph. The complexity is $O(\max(k^3/c,\log k) + \max(k^2/c,k))$. In the best case, the complexity is reduced to $O(\log k + k)$.

The serial time complexity of SVD is $O(n^2 k)$ but with randomized dimensionality reduction~\citep{gittens2013revisiting} and parallelization~\citep{constantine2011tall}, this is significantly reduced.
\subsubsection{STGD}
In STGD, we perform implicit stochastic updates, consisting of a constant number of matrix-matrix and matrix-vector products, on the set of eigenvectors and whitened samples which is of size $k \times k$. When $c \in [1, k^3 / \log k]$, we obtain a running time of $O({k^3/c})$ for computing inner products in parallel with $c$ compute cores since each core can perform an inner product to compute an element in the resulting matrix independent of other cores in linear time. For $c \in (k^3 / \log k, \infty]$, using Lemma~\ref{lem:add}, we obtain a running time of $O(\log k)$. Note that the STGD time complexity is calculated per iteration.
\subsubsection{Post-processing}
Finally, post-processing consists of sparse matrix products as well. Similar to pre-processing, this consists of multiplications involving the sparse matrices. Given $s$ number of non-zeros per column of an $O(n) \times O(k)$ matrix, the effective number of elements reduces to $O(sk)$. Hence, given $c \in [1, nks / \log s]$ cores, we need $O({nsk/c})$ time to perform the inner products for each entry of the resultant matrix. For $c \in (nks / \log s, \infty]$, using Lemma~\ref{lem:add}, we obtain a running time of $O(\log s)$.

\bigskip
Note that $nk^2$ is the complexity of computing the exact SVD and we reduce it to $O(k)$ when there are sufficient cores available. This is meant for the setting where $k$ is small.
This $k^3$ complexity of SVD on $O(k\times k)$ matrix  can be reduced to $O(k)$  using distributed SVD algorithms~e.g.~\citep{kannan2014principal,feldman2013turning}.
We note that the variational inference algorithm complexity, by Gopalan and Blei~\citep{gopalan2013efficient},  is $O(mk)$ for each iteration, where $m$ denotes the number of edges in the graph, and  $n< m < n^2$.  In the regime that $n\gg k$, our algorithm is more efficient. Moreover, a big difference is in the scaling with respect to the size of the network and ease of parallelization of our method compared to variational one.

\section{Validation methods}\label{sec:val_meth}
\subsection{$p$-value testing: }\label{sec:apdx_pval}
\begin{figure}[hbtp]
   \centering
   \psfrag{Matched}[c]{ }
   \psfrag{\$\\Pi\_\{1\}\$}[c]{\textcolor[rgb]{1,1,1}{\tiny{ $\Pi_{1}$}}}
   \psfrag{\$\\Pi\_\{2\}\$}[c]{\textcolor[rgb]{1,1,1}{\tiny{ $\Pi_{2}$}}}
   \psfrag{\$\\Pi\_\{3\}\$}[c]{\textcolor[rgb]{1,1,1}{\tiny{ $\Pi_{3}$}}}
   \psfrag{\$\\Pi\_\{4\}\$}[c]{\textcolor[rgb]{1,1,1}{\tiny{ $\Pi_{4}$}}}
   \psfrag{\$\\hat\{\\Pi\}\_\{1\}\$}[l]{\textcolor[rgb]{1,1,1}{\tiny{$\widehat{\Pi}_{1}$}}}
   \psfrag{\$\\hat\{\\Pi\}\_\{2\}\$}[l]{\textcolor[rgb]{1,1,1}{\tiny{$\widehat{\Pi}_{2}$}}}
   \psfrag{\$\\hat\{\\Pi\}\_\{3\}\$}[l]{\textcolor[rgb]{1,1,1}{\tiny{$\widehat{\Pi}_{3}$}}}
   \psfrag{\$\\hat\{\\Pi\}\_\{4\}\$}[l]{\textcolor[rgb]{1,1,1}{\tiny{$\widehat{\Pi}_{4}$}}}
   \psfrag{\$\\hat\{\\Pi\}\_\{5\}\$}[l]{\textcolor[rgb]{1,1,1}{\tiny{$\widehat{\Pi}_{5}$}}}
   \psfrag{\$\\hat\{\\Pi\}\_\{6\}\$}[l]{\textcolor[rgb]{1,1,1}{\tiny{$\widehat{\Pi}_{6}$}}}
    \includegraphics[width=1.2in,height=2in]{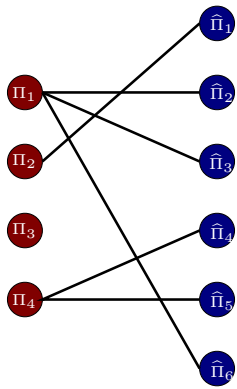}
   \caption{Bipartite graph $G_{\{\Pvalue\}}$ induced by $p$-value testing. Edges represent statistically significant relationships between ground truth and estimated communities.}
      \label{fig:match}
\end{figure}
We recover the estimated community membership matrix $\widehat{\Pi}\in \mathbb{R}^{\widehat{k}\times n}$, where $\widehat{k}$ is the number of communities specified to our method. Recall that the true community membership matrix is $\Pi$, and we consider datasets where ground truth is available. Let $i$-th row of $\widehat{\Pi}$ be denoted by $\widehat{\Pi}_i$. Our community detection method is unsupervised, which inevitably results in row permutations between $\Pi$ and $\widehat{\Pi}$ and $\widehat{k}$ may not be the same as $k$. To validate the results, we need to find a good match between the rows of $\widehat{\Pi}$ and $\Pi$. We use the notion of $p$-values 
 to test for statistically significant dependencies among a set of random variables.
The $p$-value denotes the probability of not rejecting the null hypothesis that the random variables under consideration are independent and we use the  Student's\footnote{Note that Student's $t$-test is robust to the presence of unequal variances when the sample sizes of the two are equal which is true in our setting.} $t$-test statistic~\citep{fadem2012high} to compute the $p$-value. We use multiple hypothesis testing for different pairs of estimated and ground-truth communities $\widehat{\Pi}_i, \Pi_j$ and  adjust the $p$-values to ensure a small enough false discovery rate (FDR)~\citep{strimmer2008fdrtool}.

The test statistic used for the $p$-value testing of the estimated communities is
\[
T_{ij} : = \frac{\rho \left(\widehat{\Pi}_i,\Pi_j\right)\sqrt{n-2}}{\sqrt{1-\rho \left(\widehat{\Pi}_i,\Pi_j\right)^2}}.
\]
The right $p$-value is obtained via the probability of obtaining a value (say $t_{ij}$) greater than the test statistic $T_{ij}$, and it is defined as
\[
\Pvalue(\Pi_i , \widehat{\Pi}_j):= 1- \mathbb{P}\left( t_{ij} > T_{ij}\right).
\]
Note that $T_{ij}$ has Student's $t$-distribution with degree of freedom $n-2$ (i.e. $T_{ij}\sim t_{n-2} $).  Thus, we obtain the right $p$-value\footnote{The right $p$-value accounts for the fact that when two communities are anti-correlated they are not paired up. Hence note that in the special case of block model in which the estimated communities are just permuted version of the ground truth communities, the pairing results in a perfect matching accurately.}.

In this way, we compute the $\Pvaluem$ matrix as\[
\Pvaluem(i,j):=\Pvalue\left[\widehat{\Pi}_i,\Pi_j\right], \forall i\in [k] \text{ and } j \in [\widehat{k}].
\]

\subsection{Evaluation metrics}\label{sec:defineourscores}
\paragraph{Recovery ratio: }Validating the results requires a matching of the true membership $\Pi$ with estimated membership $\widehat{\Pi}$.
Let $\Pvalue(\Pi_i , \widehat{\Pi}_j)$ denote the right $p$-value under the null  hypothesis that $\Pi_i$ and $\widehat{\Pi}_j$ are statistically independent. We use the $p$-value test to find out pairs $\Pi_i , \widehat{\Pi}_j$ which pass a specified $p$-value threshold, and we denote such pairs using  a bipartite graph $G_{\{\Pvalue\}}$. Thus, $G_{\{\Pvalue\}}$ is defined as
\[
G_{\{\Pvalue\}}:=\left(\left\{V^{(1)}_{\{\Pvalue\}},V^{(2)}_{\{\Pvalue\}}\right\}, E_{\{\Pvalue\}} \right),
\]
where the nodes in the two node sets are
\begin{align*}
&V^{(1)}_{\{\Pvalue\}}=\left\{\Pi_1,\ldots,\Pi_k\right\},\quad \\
&V^{(2)}_{\{\Pvalue\}}=\left\{\widehat{\Pi}_1,\ldots,\widehat{\Pi}_{\widehat{k}}\right\}
\end{align*}
 and the edges of $G_{\{\Pvalue\}}$ satisfy
\[(i,j)\in
E_{\{\Pvalue\}} \text{ s.t. } \Pvalue\left[\widehat{\Pi}_i,\Pi_j\right] \le 0.01
.
\]

A simple example is shown in Figure~\ref{fig:match}, in which $\Pi_{2}$  has statistically significant dependence with $\widehat{\Pi}_1$, i.e., the probability of not rejecting the null hypothesis is small (recall that null hypothesis is that they are independent). If no estimated membership vector has a significant overlap with $\Pi_3$, then $\Pi_3$ is not recovered. There can also be multiple pairings such as for $\Pi_{1}$ and $\{\widehat{\Pi}_2,\widehat{\Pi}_3,\widehat{\Pi}_6\}$. The $p$-value test between $\Pi_{1}$ and $\{\widehat{\Pi}_2,\widehat{\Pi}_3,\widehat{\Pi}_6\}$ indicates that probability of not rejecting the null hypothesis is small, i.e., they are independent. We use $0.01$ as the threshold. The same holds for $\Pi_{2}$ and $\{\widehat{\Pi}_1\}$ and for $\Pi_{4}$ and $\{\widehat{\Pi}_4,\widehat{\Pi}_5\}$. There can be a perfect one to one matching like for $\Pi_{2}$ and $\widehat{\Pi}_1$ as well as a multiple matching such as for $\Pi_{1}$ and $\{\widehat{\Pi}_2,\widehat{\Pi}_3,\widehat{\Pi}_6\}$. Or another multiple matching such as for $\{\Pi_{1},\Pi_{2}\} $ and $\widehat{\Pi}_3$.

Let $\degree_i$ denote the degree of ground truth community $i\in[k]$ in $G_{\{\Pvalue\}}$, we define the recovery ratio as follows.
\begin{definition}
The \emph{recovery ratio} is defined as
\begin{equation*}
\mathcal{R}:=\frac{1}{k} \sum\limits_{i}\mathbb{I}\left\{\degree_i >0 \right\}, \quad i\in[k]
  \end{equation*}
    where $\mathbb{I}(x)$ is the indicator function whose value equals one if $x$ is true.
\end{definition}
The perfect case is that all the memberships have at least one significant overlapping estimated membership, giving a recovery ratio of $100\%$.
\paragraph{Error function: }For performance analysis of our learning algorithm, we use an error function given as follows:
\begin{definition}
The average error function is defined as

\[
\mathcal{E}: =\frac{1}{k}\sum\limits_{
      (i,j)\in E_{\{\Pvalue\}}
}
 \left\{\frac{1}{n}\sum\limits_{x\in |X|} {\llvert \widehat{\Pi}_i (x)- \Pi_j(x) \llvert}\right\},
\] where $E_{\{\Pvalue\}}
$ denotes the set of edges based on thresholding of the $p$-values.
\end{definition}

The error function incorporates two aspects, namely the $l_1$ norm error between each estimated community  and the corresponding paired ground truth community, and the error induced by false pairings between the estimated and ground-truth communities through $p$-value testing. For the former $l_1$ norm error, we normalize with $n$ which is reasonable and results in the range of the error in $[0,1]$. For the latter, we define the average error function as the summation of all paired memberships errors divided by the true number of communities $k$. In this way we penalize falsely discovered pairings by summing them up. Our error function can be greater than 1 if there are too many falsely discovered pairings through $p$-value testing (which can be as large as $k\times \widehat{k}$).

\paragraph{Bridgeness: }
Bridgeness in overlapping communities is an interesting measure to evaluate.  A bridge is defined as a vertex that crosses structural holes between discrete groups of people and bridgeness analyzes the extent to which a given vertex is shared among different communities~\citep{nepusz2008fuzzy}. Formally, the bridgeness of a vertex $i$ is defined as 
\begin{equation}\label{eq:bridgeness}
b_i:=1-\sqrt{\frac{\widehat{k}}{\widehat{k}-1}\sum\limits_{j=1}^{\widehat{k}}{\left(\widehat{\Pi}_i(j)-\frac{1}{\widehat{k}}\right)}^2}.
\end{equation}
Note that centrality measures should be used in conjunction with bridge score to distinguish outliers from genuine bridge nodes~\citep{nepusz2008fuzzy}.
The \emph{degree-corrected bridgeness} is  used to evaluate our results and is defined as
\begin{equation}\label{eq:degreebridgeness}
\mathcal{B}_i:=D_ib_i,
\end{equation}
where $D_i$ is  degree of node $i$.

\section{Experimental Results}
\label{sec:results}
The specifications of the machine on which we run our code are given in Table~\ref{tab:sys_spec}.


\paragraph{Results on Synthetic Datasets: }

\begin{table}
   \centering
   \small \begin{tabular}{@{} |l|l| @{}}
\hline
Hardware / software & Version\\
\hline
\hline
CPU & Dual 8-core Xeon $@$ 2.0GHz\\
Memory & 64GB DDR3\\
GPU & Nvidia Quadro K5000\\
CUDA Cores & 1536\\
Global memory & 4GB GDDR5\\
CentOS & Release 6.4 (Final)\\
GCC & 4.4.7 \\
CUDA & Release 5.0\\
CULA-Dense & R16a\\
\hline
   \end{tabular}
   \caption{System specifications.}
   \label{tab:sys_spec}
\end{table}


We perform experiments for both the stochastic block model ($\alpha_0=0$) and the mixed membership model. For the mixed membership model, we set the concentration parameter $\alpha_0 =1$. We note that the error is around $8\% - 14\%$ and the running times are under a minute, when $n \leq 10000$ and $n \gg k$\footnote{The code is available at \url{https://github.com/FurongHuang/Fast-Detection-of-Overlapping-Communities-via-Online-Tensor-Methods}}.

We observe that more samples result in a more accurate recovery of memberships which matches intuition and theory. Overall, our learning algorithm performs better in the stochastic block model case than in the mixed membership model case although we note that the accuracy is quite high for practical purposes. Theoretically, this is expected since smaller concentration parameter $\alpha_0$ is easier for our algorithm to learn~\citep{AnandkumarEtal:community12COLT}. Also, our algorithm is scalable to an order of magnitude more in $n$ as illustrated by experiments on real-world large-scale datasets.

Note that we threshold the estimated memberships to clean the results. There is a tradeoff between match ratio and average error via different thresholds. In synthetic experiments, the tradeoff is not evident since a perfect matching is always present. However, we need to carefully handle this in experiments involving real data.

\paragraph{Results on Topic Modeling:}

We perform experiments for the bag of words dataset~\citep{Bache+Lichman:2013} for The New York Times. We set the concentration parameter to be $\alpha_0=1$ and observe top recovered words in numerous topics. The results are in Table~\ref{tab:Nytimes}. Many of the results are expected. For example, the top words in topic \# 11 are all related to some bad personality.

We also present the words with most spread membership, i.e., words that belong to many topics as in Table~\ref{tab:nytimesbridge}. As expected, we see minutes, consumer, human, member and so on. These words can appear in a lot of topics, and we expect them to connect topics.

\tiny
\begin{table}[htbp]
\scriptsize
   \centering
   \begin{tabular}{@{} |l|lllll|@{}} 
\hline
Topic \#&  & Top Words& & &  \\
\hline
\hline
1& prompting	&	complicated	&	eviscerated	&	predetermined	&	lap\\
	&	renegotiating	&	loose	&	entity	&	legalese	&	justice\\
	\hline	
2 & hamstrung	&	airbrushed	&	quasi	&	outsold	&	fargo	\\
&	ennobled	&	tantalize	&	irrelevance	&	noncontroversial	&	untalented	\\	
\hline
3 &scariest	&	pest	&	knowingly	&	causing	&	flub\\
	&	mesmerize	&	dawned	&	millennium	&	ecological	&	ecologist	\\
	\hline	
4 & reelection	&	quixotic	&	arthroscopic	&	versatility	&	commanded\\
	&	hyperextended	&	anus	&	precipitating	&	underhand	&	knee\\
	\hline
5 &believe	&	signing	&	ballcarrier	&	parallel	&	anomalies\\
	&	munching	&	prorated	&	unsettle	&	linebacking	&	bonus\\
	\hline
6 &gainfully	&	settles	&	narrator	&	considerable	&	articles\\
	&	narrative	&	rosier	&	deviating	&	protagonist	&	deductible\\
	\hline
7 &faithful	&	betcha	&	corrupted	&	inept	&	retrench \\
	&	martialed	&	winston	&	dowdy	&	islamic	&	corrupting	\\
	\hline
8 &capable	&	misdeed	&	dashboard	&	navigation	&	opportunistically\\
	&	aerodynamic	&	airbag	&	system	&	braking	&	mph\\
	\hline
9 &apostles	&	oracles	&	believer	&	deliberately	&	loafer	\\
&	gospel	&	apt	&	mobbed	&	manipulate	&	dialogue\\
\hline
10 & physique	&	jumping	&	visualizing	&	hedgehog	&	zeitgeist	\\
&	belonged	&	loo	&	mauling	&	postproduction	&	plunk\\
\hline
11 &smirky	&	silly	&	bad	&	natured	&	frat	\\
&	thoughtful	&	freaked	&	moron	&	obtuse	&	stink	\\
\hline
12 &offsetting	&	preparing	&	acknowledgment	&	agree	&	misstating\\
	&	litigator	&	prevented	&	revoked	&	preseason	&	entomology\\
	\hline
13 &undertaken	&	wilsonian	&	idealism	&	brethren	&	writeoff	\\
&	multipolar	&	hegemonist	&	multilateral	&	enlargement	&	mutating	\\	
\hline
14 & athletically	&	fictitious	&	myer	&	majorleaguebaseball	&	familiarizing\\
	&	resurrect	&	slug	&	backslide	&	superseding	&	artistically	\\
	\hline
15 & dialog	&	files	&	diabolical	&	lion	&	town	\\
&	password	&	list	&	swiss	&	coldblooded	&	outgained	\\
\hline
16 & recessed	&	phased	&	butyl	&	lowlight	&	balmy\\
	&	redlining	&	prescription	&	marched	&	mischaracterization	&	tertiary\\
	\hline
17 & sponsor	&	televise	&	sponsorship	&	festival	&	sullied\\
	&	ratification	&	insinuating	&	warhead	&	staged	&	reconstruct	\\
	\hline
18 &trespasses	&	buckle	&	divestment	&	schoolchild	&	refuel	\\
&	ineffectiveness	&	coexisted	&	repentance	&	divvying	&	overexposed	\\	
\hline
 \end{tabular}
   \caption{Top recovered topic groups from the New York Times dataset along with the words present in them.}
   \label{tab:Nytimes}
\end{table}
\normalsize

\begin{table}[htbp]
   \centering
   \begin{tabular}{@{} |l| @{}} 
\hline
Keywords \\
\hline
      \hline
   minutes, 
consumer, 
human, 
member, 
friend, 
program, 
board, 
cell, 
insurance, 
shot	\\
\hline
   \end{tabular}
   \caption{The top ten words which occur in multiple contexts in the New York Times dataset.}
   \label{tab:nytimesbridge}
\end{table}

\paragraph{Results on Real-world Graph Datasets: }We describe the results on real datasets summarized in Table~\ref{tab:data_info} in detail below. The simulations are summarized in Table~\ref{table:businessresults}.

\begin{table}[htbp]
\small
   \centering
      \begin{tabular}{@{} |l|l|l|l|l| @{}} 
\hline
Statistics				& Facebook 	& Yelp  			& DBLP sub & DBLP \\
\hline
\hline
$\lvert E \rvert$		&	766,800	&672,515			&5,066,510 	& 16,221,000\\
$\lvert V\rvert$		&	18,163   	&10,010$+$28,588	& 116,317		& 1,054,066\\
GD 				& 0.004649 	& 0.000903 		& 0.000749 	& 0.000029\\
$k$				&	360		&159				&250 		& 6,003\\
AB                  		& 0.5379           &  0.4281      		& 0.3779 		& 0.2066\\
ADCB			&	47.01	&30.75			&48.41 		& 6.36\\
\hline
   \end{tabular}
   \caption{Summary of real datasets used in our paper: $\lvert V\rvert$	is the number of nodes in the graph, $\lvert E \rvert$ is the number of edges, GD is the graph density given by $\frac{2\lvert E\lvert}{\lvert V\rvert\left(\lvert V\rvert-1\right)}$,  
   $k$ is the number of communities,
   AB is the average bridgeness and ADCB is the average degree-corrected bridgeness(explained in Section~\ref{sec:val_meth}).}
 \label{tab:data_info}
\end{table}

\begin{table}[h]
\centering
{\scriptsize
\begin{tabular}{@{} |l|l|l|l|l|l|l| @{}}
\hline
Data 	& Method 		& $\widehat{k}$ 	& Thre 	& $\mathcal{E}$ 	&$\mathcal{R} (\%)$		&  Time(s)  \\
\hline
 \hline
		&Ten(sparse) 		&$10$		&$0.10$  		& $0.063$     	& $13$   		&$35$  \\
		&Ten(sparse) 		&$100$		&$0.08$ 		&$0.024$ 		& $62$  		&$309$ \\
		&Ten(sparse) 		&$100$		&$0.05$ 		& $0.118$ 	& $95$ 		&$309$ \\
 		&Ten(dense)  		&$100$		&$0.100$  	& $0.012$     	& $39$ 		 &$190$  \\
		&Ten(dense)  	 	&$100$		&$0.070$ 		& $0.019$ 	& $100$ 	 	&$190$ \\
FB		& Variational 		&$100$ 		&--			& $0.070$     	& $100$ 	 	&$10,795$  \\
		& Ten(dense)  		&$500$		&$0.020$  	& $0.014$  	& $71$ 		&$468$  \\
		& Ten(dense) 		&$500$		&$0.015$  	& $0.018$ 	& $100$		&$468$ \\
		& Variational 			&$500$    		&--			& $0.031$  	& $100$	 	&$86,808$  \\

 \hline
		&Ten(sparse) 		&$10$    		&$0.10$  		&$0.271$     	&$43$ 		&$10$  \\
		&Ten(sparse) 		&$100$ 		&$0.08$ 		&$0.046$ 		&$86$ 		&$287$ \\
		&Ten(dense) 		&$100$    		& $0.100$  	& $0.023$     	&  $43$ 		&$1,127$  \\
YP		&Ten(dense) 		&$100$ 		&$0.090$ 		& $ 0.061$ 	& $80$ 		&$1,127$ \\
		&Ten(dense)		&$500$    		&$0.020$   	& $0.064$  	& $72$		&$1,706$  \\
		&Ten(dense)		&$500$ 		&$0.015$  	& $0.336$ 	& $100$		&$1,706$ \\
\hline
		&Ten(dense) 		&$100$ 		&$0.15$  		&$0.072$ 		&$36$ 		&$7,664$ \\
		&Ten(dense) 		&$100$    		&$0.09$  		&$0.260$     	&$80$ 		&$7,664$  \\
			&Variational				&$100$		&--			&$7.453$		&$99$		&$69,156$\\
DB sub		&Ten(dense)  		&$500$    		&$0.10$  		&$0.010$  	&$19$		&$10,157$  \\
			&Ten(dense) 		&$500$ 		&$0.04$   		&$0.139$ 		&$89$		&$10,157$ \\
			&Variational				&$500$		&--			&$16.38$		&$99$		&$558,723$\\
\hline
		&Ten(sparse)  		&$10$    		&$0.30$ 		&$0.103$  	&$73$     		&$4716$  \\
DB		&Ten(sparse) 		&$100$ 		&$0.08$  		&$0.003$   	&$57$ 		&$5407$ \\
		&Ten(sparse) 		&$100$ 		&$0.05$  		&$0.105$ 	       	&$95$ 		&$5407$ \\
\hline
   \end{tabular}
   }
\caption{Yelp, Facebook and DBLP main quantitative evaluation of the tensor method versus the variational method:
 $\widehat{k}$ is the community number specified to our algorithm, Thre is the threshold for picking significant estimated membership entries. Refer to Table~\ref{tab:data_info} for statistics of the datasets.
 }
\label{table:businessresults}
\end{table}

\begin{table}[htbp]
   \scriptsize
   \centering
   \begin{tabular}{@{} |l|l|l|@{}} 
\hline
      \textbf{Business} & \textbf{RC} & \textbf{Categories}\\
\hline
      \hline
\textbf{Four Peaks Brewing Co}		&735			& Restaurants, Bars, American (New), Nightlife, Food, Pubs, Tempe\\
\textbf{Pizzeria Bianco}			&803			& Restaurants, Pizza,Phoenix\\
\textbf{FEZ}					&652			& Restaurants, Bars, American (New), Nightlife, Mediterranean, Lounges\\
 & & Phoenix\\
\textbf{Matt's Big Breakfast}		&689			& Restaurants, Phoenix, Breakfast\& Brunch\\
\textbf{Cornish Pasty Company}	&580			& Restaurants, Bars, Nightlife, Pubs, Tempe\\
\textbf{Postino Arcadia}			&575			& Restaurants, Italian, Wine Bars, Bars, Nightlife, Phoenix\\
\textbf{Cibo}					&594			& Restaurants, Italian, Pizza, Sandwiches, Phoenix\\
\textbf{Phoenix Airport}			&862			&  Hotels \& Travel, Phoenix\\
\textbf{Gallo Blanco Cafe}			&549			& Restaurants, Mexican, Phoenix\\
\textbf{The Parlor}				&489			& Restaurants, Italian, Pizza, Phoenix\\
\hline
   \end{tabular}
   \caption{Top 10 bridging businesses in   Yelp    and categories they belong to. ``RC'' denotes review counts for that particular business.}
   \label{tab:bridgeYELPwithCom}
\end{table}





The results are presented in Table~\ref{table:businessresults}. We note that our method, in both dense and sparse implementations, performs very well compared to the state-of-the-art variational method.
For the Yelp dataset, we have a bipartite graph where the business nodes are on one side and user nodes on the other and use the review stars as the edge weights.  
In this bipartite setting, 
the variational code provided by Gopalan et al~\citep{gopalan2012scalable} does not work on since it is not applicable to non-homophilic models. Our approach does not have this restriction. Note that we use our dense implementation on the GPU to run experiments with large number of communities $k$ as the device implementation is much faster in terms of running time of the STGD step.
On the other hand, the sparse implementation on CPU is fast and memory efficient in the case of sparse graphs with a small number of communities while the dense implementation on GPU is faster for denser graphs such as Facebook. Note that data reading time for DBLP is around 4700 seconds, which is not negligible as compared to other datasets (usually within a few seconds). Effectively, our algorithm, excluding the file I/O time, executes within two minutes for $k=10$ and within ten minutes for $k=100$.

\begin{figure}[h]
\psfrag{Number of categories}[l]{\tiny{Business Category ID}}
\psfrag{Number of business}[l]{\tiny{\# business $\ $}}
\psfrag{Distribution of Categories: }[c]{\scriptsize{ }}
 \includegraphics[width=0.49\columnwidth]{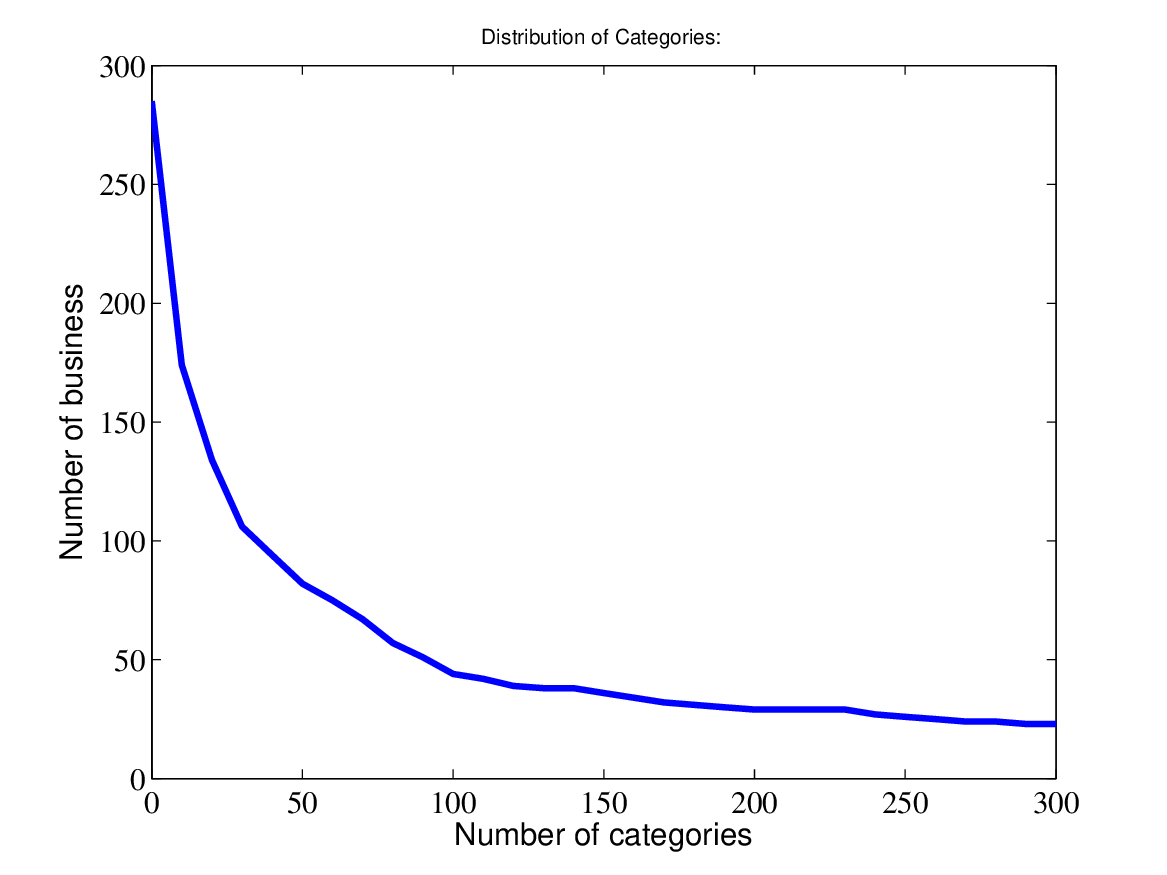}
\psfrag{match ratio}[l]{\tiny{Recovery Ratio}}
\psfrag{average error}[l]{\tiny{Average Error $\quad \quad $}}
\includegraphics[width=0.49\columnwidth]{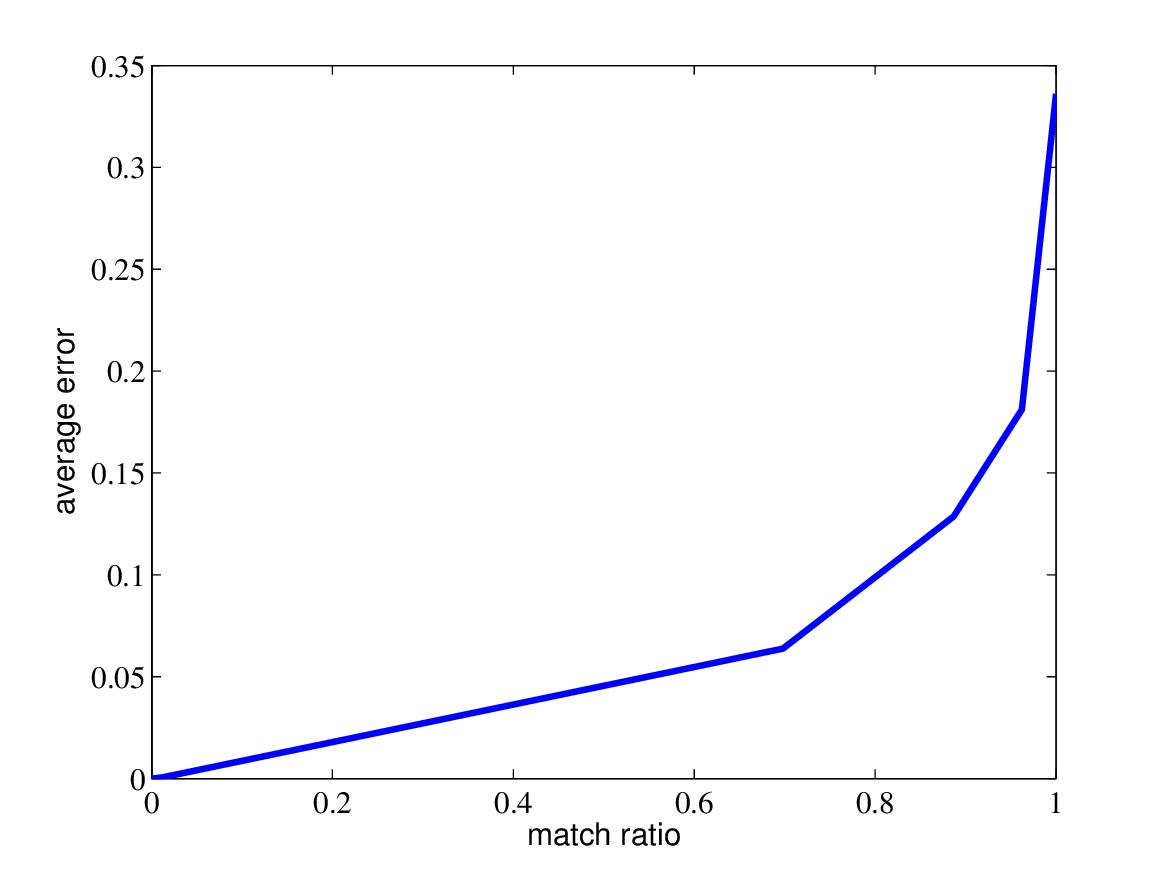}
\caption{Distribution of business categories (left) and result tradeoff between recovery ratio and error for yelp (right).}
\label{fig:tradeoff}
\end{figure}

\paragraph{Interpretation on Yelp Dataset: }The ground truth on business attributes such as location and type of business are available (but not provided to our algorithm) and we provide the distribution in Figure~\ref{fig:tradeoff} on the left side. There is also a natural trade-off between recovery ratio and average error or between attempting to recover all the business communities and the accuracy of recovery. We can either recover top significant communities with high accuracy or recover more with lower accuracy. We demonstrate the trade-off in Figure~\ref{fig:tradeoff} on the right side.

We select the top ten categories recovered with the lowest error and report the business with  highest weights in $\widehat{\Pi}$. Among the matched communities, we find the business with the highest membership weight (Table~\ref{tab:topbusinesses}). We can see that most of the ``top'' recovered businesses are  rated   high. 
Many of the categories in the top ten list are restaurants as they have a large number of reviewers. Our method can recover restaurant category with high accuracy, and the specific  restaurant in the category is a popular result (with high number of stars). Also, our method can also recover many of the categories with low review counts accurately like hobby shops, yoga, churches, galleries and religious organizations which are the ``niche'' categories with a dedicated set of reviewers, who mostly do not review other categories. 

\begin{table}[htbp]
  \scriptsize
  \centering
   \begin{tabular}{@{} |l|l|l|c|c|c|c|c|c| @{}}
\hline
      Category & Business & Star(B) & Star(C) &
      RC(B) & RC(C)\\
\hline
      Latin American & Salvadoreno & $4.0$ & $3.94$	
      &$36$ 	&$93.8$\\
      Gluten Free & P.F. Chang's & $3.5$ & $3.72$		
      &$55$	&$50.6$\\
      Hobby Shops & Make Meaning & $4.5$ & $4.13$			
      &$14$	&$7.6$\\
      Mass Media & KJZZ $91.5$FM & $4.0$ & $3.63$			
      &$13$	&$5.6$\\
      Yoga & Sutra Midtown & $4.5$ & $4.55$					
      &$31$	&$12.6$\\
      Churches & St Andrew Church & $4.5$ & $4.52$
      &$3$	&$4.2$\\
      Art Galleries &Sette Lisa &$4.5$ & $4.48$         		
      &$4$		&$6.6$\\
      Libraries & Cholla Branch & $4.0$ & $4.00$			
      &$5$		&$11.2$\\
      Religious & St Andrew Church & $4.5$ &$4.40$	
      &$3$ &$4.2$\\
      Wickenburg & Taste of Caribbean & $4.0$ & $3.66$
      &$60$	& $6.7$\\
\hline
   \end{tabular}
   \caption{Most accurately recovered categories and businesses with highest membership weights for the Yelp dataset. ``Star(B)'' denotes the review stars that the business receive and ``Star(C)'', the average review stars that businesses in that category receive.  ``RC(B)'' denotes the review counts for that business and ``RC(C)'' , the average review counts in that category. }
   \label{tab:topbusinesses}
\end{table}

The top bridging nodes recovered by our method for the Yelp dataset are given in the    Table~\ref{tab:bridgeYELPwithCom}.  The bridging nodes have multiple attributes typically, the type of business and its location. In addition, the categories may also be hierarchical: within restaurants, different cuisines such as Italian, American or Pizza are recovered by our method. Moreover, restaurants which also function as bars or lounges are also recovered as top bridging nodes in our method. Thus, our method can recover multiple attributes for the businesses efficiently.

Among all $11537$ businesses, there are $89.39\%$ of them are still open. We only select those businesses which are still open. There are $285$ categories in total. After we remove all the categories having no more than $20$ businesses within it, there are $134$ categories that remain.  We generate community membership matrix for business categories $\Pi_c \in \mathbb{R}^{k_c \times n}$ where $k_c:=134$ is the number of remaining categories and $n:=10141$ is the number of business remaining after removing all the negligible categories. All the businesses collected in the Yelp data are in AZ except 3 of them (one is in CA, one in CO and the other in SC). We remove the three businesses outside AZ. We notice that most of the businesses are spread out in $25$ cities. Community membership matrix for location is defined as $\Pi \in \mathbb{R}^{k_l \times n}$ where $k_l:=25$ is the number cities and $n:=10010$ is number of businesses. Distribution of locations are in Table~\ref{tab:business_location}.  The stars a business receives can vary from $1$ (the lowest) to $5$ (the highest). The higher the score is, the more satisfied the customers are. The average star score is $3.6745$. The distribution is given in Table~\ref{tab:business_scores}. There are also review counts for each business which are the number of reviews that business receives from all the users. The minimum review counts is $3$ and the maximum is $862$. The mean of review counts is $20.1929$. The preprocessing helps us to pick out top communities.

There are $5$ attributes associated with all the $11537$ businesses, which are ``open'', ``Categories'', ``Location'', ``Review Counts'' and ``Stars''.  We model ground truth communities as a combination of ``Categories'' and ``Location''. We select business categories with more than $20$ members and remove all businesses which are closed. $10010$ businesses are remained.  Only $28588$ users are involved in reviews towards the $10010$ businesses. There are $3$ attributes associated with all the $28588$ users, which are ``Female'', ``Male'', ``Review Counts'' and ``Stars''. Although we do not directly know the gender information from the dataset, a name-gender guesser\footnote{ \url{https://github.com/amacinho/Name-Gender-Guesser} by Amac Herdagdelen.} is used to estimate gender information using names.

\begin{table}[htbp]
   \centering
   \begin{tabular}{@{} |l|c|r| @{}}
\hline
      Star Score    & Num of businesses & Percentage \\
\hline
\hline
      $1.0$      & $108$ & $0.94\%$ \\
       $1.5$     & $170$     &  $1.47\%$ \\
      $2.0$      & $403$  & $3.49\%$ \\
      $2.5$      & $1011$  & $8,76\%$ \\
      $3.0$ 	    & $1511$   &  $13.10\%$ \\
      $3.5$      &$2639$    & $22.87\%$\\
      $4.0$      &$2674$    & $23.18\%$\\
      $4.5$      &$1748$    & $15.15\%$\\
      $5.0$      &$1273$    & $11.03\%$\\
\hline
   \end{tabular}
   \caption{Table for distribution of business star scores.}
   \label{tab:business_scores}
\end{table}

\begin{table}[htbp]
 \scriptsize  \centering
   \begin{tabular}{@{} |l|c|r| @{}}
\hline
      City    & State & Num of business \\
\hline
\hline
 Anthem  & AZ&	34\\
 Apache Junction  & AZ&	46\\
 Avondale & AZ	&	129\\
 Buckeye & AZ	&	31\\
 Casa Grande & AZ &		48\\
 Cave Creek & AZ	 &	65\\
 Chandler 	& AZ	 & 865\\
 El Mirage 	& AZ	&11\\
 Fountain Hills 	& AZ	& 49\\
 Gilbert 	& AZ	& 439\\
 Glendale 	& AZ	& 611\\
 Goodyear 	& AZ	&126\\
 Laveen 	& AZ	& 22\\
 Maricopa 	& AZ	& 31\\
 Mesa 	& AZ	& 898\\
 Paradise Valley 	& AZ	& 57\\
 Peoria 	& AZ	& 267\\
 Phoenix 	& AZ	& 4155\\
 Queen Creek 	& AZ	& 78\\
 Scottsdale 	& AZ	& 2026\\
 Sun City 	& AZ	& 37\\
 Surprise 	& AZ	& 161\\
 Tempe 	& AZ	& 1153\\
 Tolleson 	& AZ	& 22\\
 Wickenburg 	& AZ	& 28\\
\hline
   \end{tabular}
   \caption{Distribution of business locations. Only top cities with more than $10$ businesses are presented.}
   \label{tab:business_location}
\end{table}



\begin{figure*}[h]
\psfrag{Female}[c]{\scriptsize{Female}}
\psfrag{Male}[c]{\scriptsize{Male}}
\psfrag{Nail Salon}[c]{\scriptsize{Nail Salon}}
\psfrag{Tires}[c]{\scriptsize{Tires}}
\def\svgwidth{3in}
\begingroup%
  \makeatletter%
  \providecommand\color[2][]{%
    \errmessage{(Inkscape) Color is used for the text in Inkscape, but the package 'color.sty' is not loaded}%
    \renewcommand\color[2][]{}%
  }%
  \providecommand\transparent[1]{%
    \errmessage{(Inkscape) Transparency is used (non-zero) for the text in Inkscape, but the package 'transparent.sty' is not loaded}%
    \renewcommand\transparent[1]{}%
  }%
  \providecommand\rotatebox[2]{#2}%
  \ifx\svgwidth\undefined%
    \setlength{\unitlength}{6616bp}%
    \ifx\svgscale\undefined%
      \relax%
    \else%
      \setlength{\unitlength}{\unitlength * \real{\svgscale}}%
    \fi%
  \else%
    \setlength{\unitlength}{\svgwidth}%
  \fi%
  \global\let\svgwidth\undefined%
  \global\let\svgscale\undefined%
  \makeatother%
  \begin{picture}(1,0.74365175)%
    \put(0,0){\includegraphics[width=\unitlength]{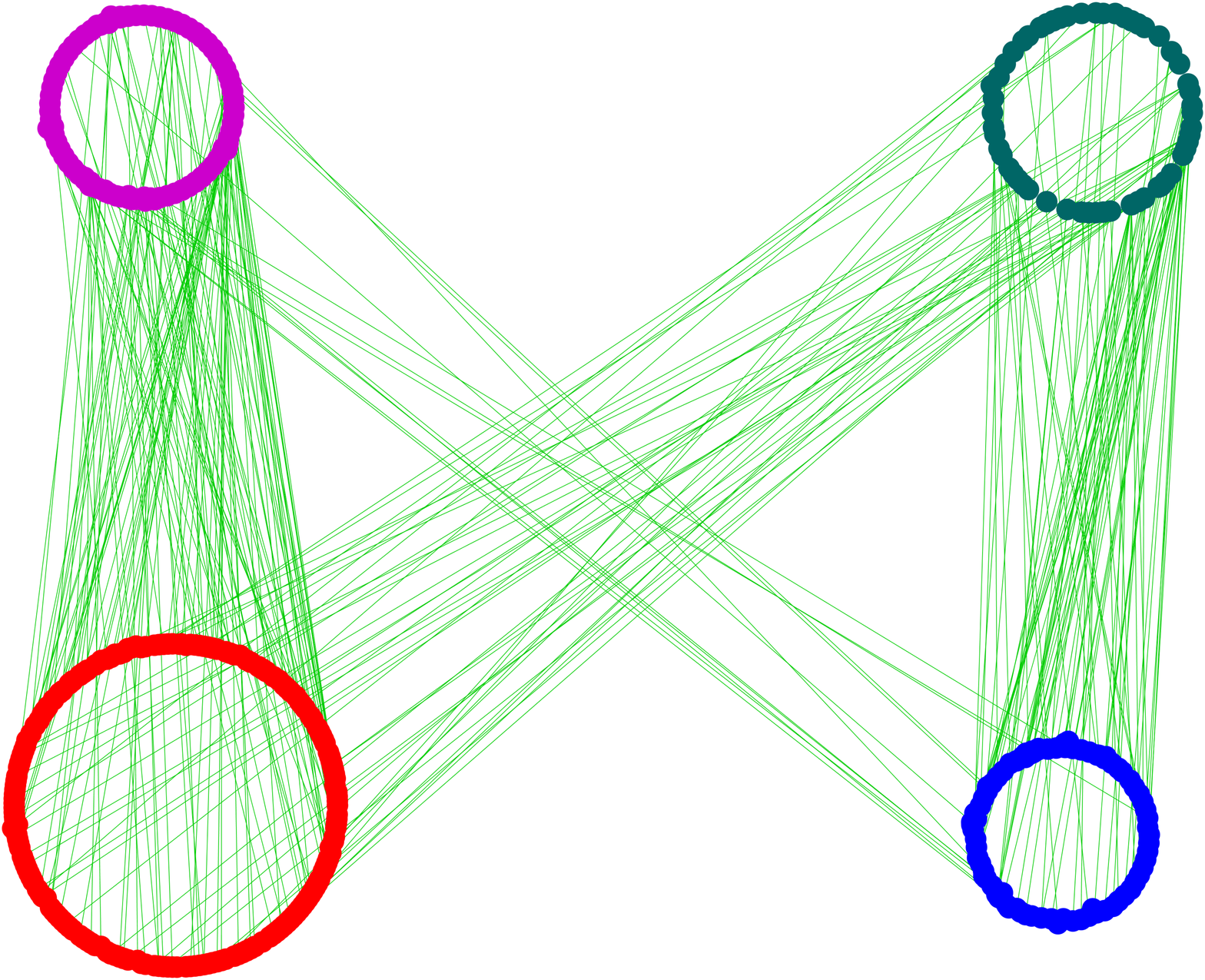}}%
    \put(0.90750797,0.7193994){\color[rgb]{0,0,0}\makebox(0,0)[lb]{\smash{Tires}}}%
    \put(0.90760897,0.02047753){\color[rgb]{0,0,0}\makebox(0,0)[lb]{\smash{Male}}}%
    \put(0.0,0.00047753){\color[rgb]{0,0,0}\makebox(0,0)[lb]{\smash{Female}}}%
    \put(0.0,0.7193994){\color[rgb]{0,0,0}\makebox(0,0)[lb]{\smash{Nail Salon}}}%
  \end{picture}%
\endgroup%
\psfrag{Female}[c]{\scriptsize{Female}}
\psfrag{Male}[c]{\scriptsize{Male}}
\psfrag{Nail Salon}[c]{\scriptsize{Nail Salon}}
\psfrag{Tires}[c]{\scriptsize{Tires}}
\hspace{0.1\textwidth}
\def\svgwidth{3in}
\begingroup%
  \makeatletter%
  \providecommand\color[2][]{%
    \errmessage{(Inkscape) Color is used for the text in Inkscape, but the package 'color.sty' is not loaded}%
    \renewcommand\color[2][]{}%
  }%
  \providecommand\transparent[1]{%
    \errmessage{(Inkscape) Transparency is used (non-zero) for the text in Inkscape, but the package 'transparent.sty' is not loaded}%
    \renewcommand\transparent[1]{}%
  }%
  \providecommand\rotatebox[2]{#2}%
  \ifx\svgwidth\undefined%
    \setlength{\unitlength}{4135.00356bp}%
    \ifx\svgscale\undefined%
      \relax%
    \else%
      \setlength{\unitlength}{\unitlength * \real{\svgscale}}%
    \fi%
  \else%
    \setlength{\unitlength}{\svgwidth}%
  \fi%
  \global\let\svgwidth\undefined%
  \global\let\svgscale\undefined%
  \makeatother%
  \begin{picture}(1,0.74365372)%
    \put(0,0){\includegraphics[width=\unitlength]{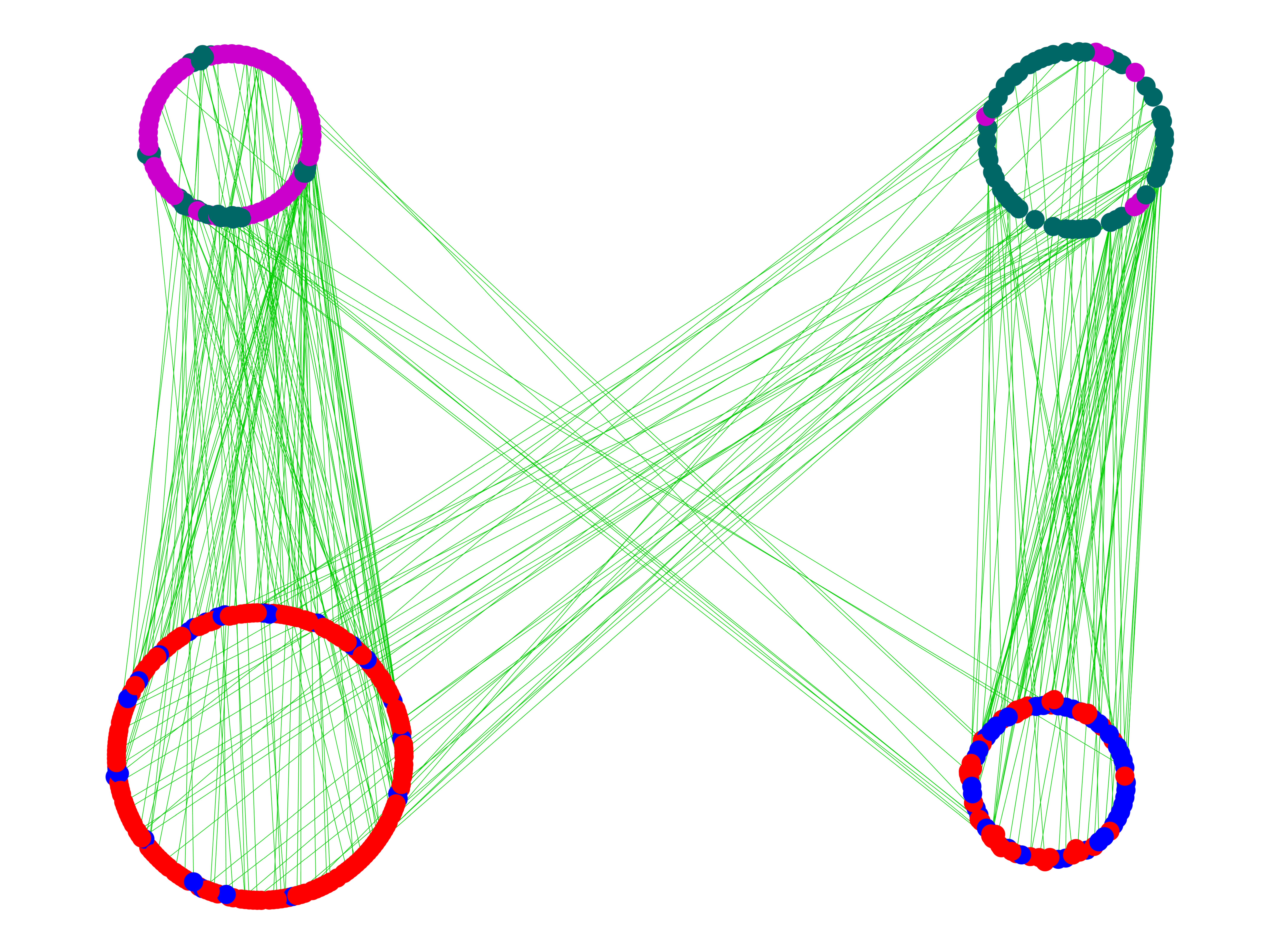}}%
    \put(0.90750797,0.7193994){\color[rgb]{0,0,0}\makebox(0,0)[lb]{\smash{Tires}}}%
    \put(0.90760897,0.02047753){\color[rgb]{0,0,0}\makebox(0,0)[lb]{\smash{Male}}}%
    \put(0.0,0.00047753){\color[rgb]{0,0,0}\makebox(0,0)[lb]{\smash{Female}}}%
    \put(0.0,0.7193994){\color[rgb]{0,0,0}\makebox(0,0)[lb]{\smash{Nail Salon}}}%
  \end{picture}%
\endgroup%
\caption{{
Ground truth (left) vs estimated business and user categories (right). The error in the estimated graph due to misclassification is shown by the mixed colours.
}}
\label{fig:visualization}
\end{figure*}

We provide some sample visualization results in Figure~\ref{fig:visualization} for both the ground truth and the estimates from our algorithm. We sub-sample  the users and businesses, group the users into male and female categories, and consider nail salon and tire businesses. Analysis of ground truth reveals that nail salon and tire businesses are very discriminative of the user genders, and thus we employ them for visualization. 
We note that both the nail salon and tire businesses are categorized with high accuracy, while users are categorized with poorer accuracy.


Our algorithm can also recover the attributes of users. However, the ground truth available about users
is far more limited than businesses, and we only have information on gender, average review counts and average stars (we infer the gender of the users through their  names). Our algorithm can  recover all these attributes. We observe that gender is the hardest to recover while review counts is the easiest.
We see that the other user attributes recovered by our algorithm correspond to valuable user information such as their interests, location, age, lifestyle, etc. This is useful, for instance, for businesses studying the characteristics of their users, for delivering better personalized advertisements for users, and so on.

\paragraph{Facebook Dataset: }A snapshot of the Facebook network of UNC~\citep{facebook} is provided with user attributes.
The ground truth communities are based on user attributes given in the dataset which are not exposed to the algorithm. There are $360$ top communities with sufficient (at least 20) users. Our algorithm can recover these attributes with high accuracy; see main paper for our method's results compared with variational inference result~\citep{gopalan2012scalable}. 

\begin{figure*}[h]
\small \centering
\psfrag{Recovery ratio vs threshold under alpha0s}[l]{\tiny{}}
\psfrag{Threshold}[c]{\small{Threshold}}
\psfrag{Rec. ratio}[c]{\small{Recovery ratio}}
\psfrag{alpha0.1}[l]{\small{$\alpha_0$:0.1}}
\psfrag{alpha0.2}[l]{\small{$\alpha_0$:0.2}}
\psfrag{alpha0.3}[l]{\small{$\alpha_0$:0.3}}
\psfrag{alpha0.4}[l]{\small{$\alpha_0$:0.4}}
\psfrag{alpha0.5}[l]{\small{$\alpha_0$:0.5}}
\psfrag{alpha0.6}[l]{\small{$\alpha_0$:0.6}}
\psfrag{alpha0.7}[l]{\small{$\alpha_0$:0.7}}
\psfrag{alpha0.8}[l]{\small{$\alpha_0$:0.8}}
\psfrag{alpha0.9}[l]{\small{$\alpha_0$:0.9}}
\includegraphics[width=0.49\textwidth]{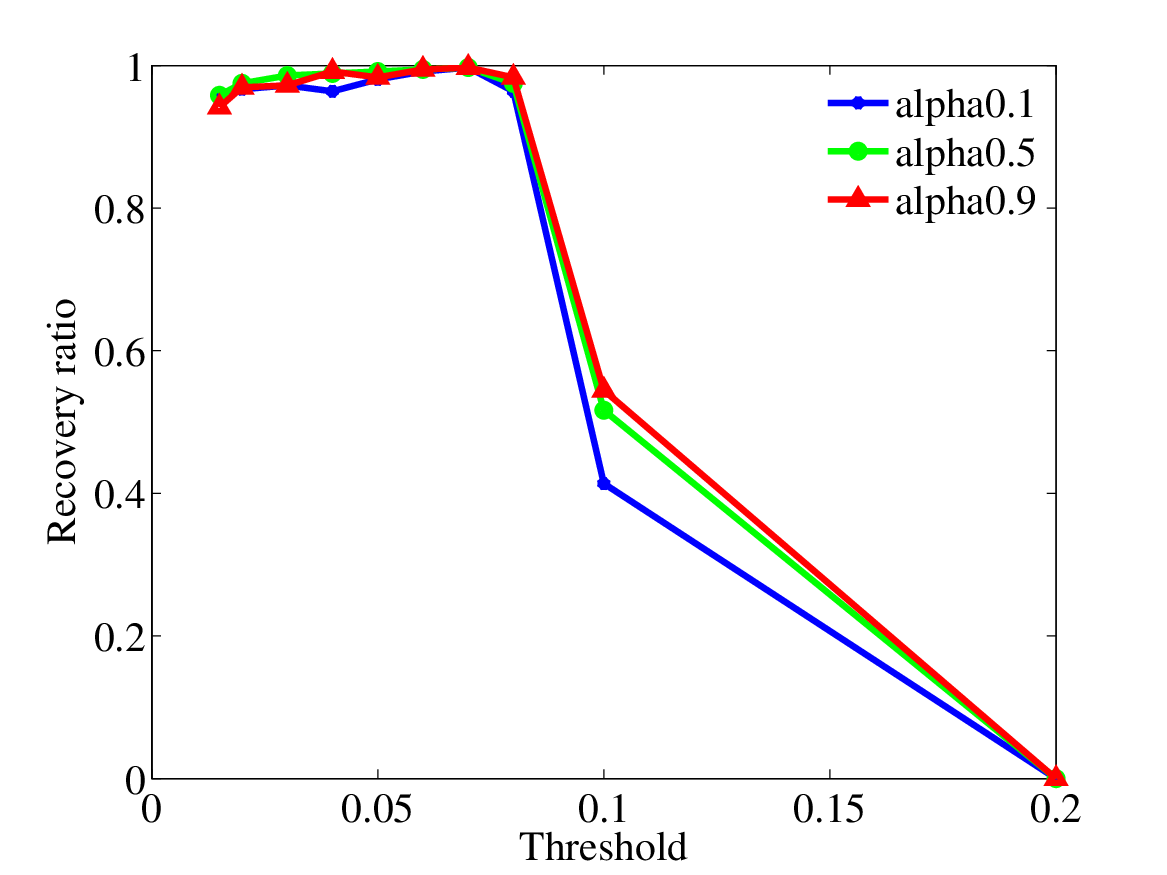}
\small \centering
\psfrag{Error vs threshold under alpha0s}[l]{\tiny{}}
\psfrag{Threshold}[c]{\small{Threshold}}
\psfrag{Error}[c]{\small{Error}}
\psfrag{alpha0.1}[l]{\small{$\alpha_0$:0.1}}
\psfrag{alpha0.2}[l]{\small{$\alpha_0$:0.2}}
\psfrag{alpha0.3}[l]{\small{$\alpha_0$:0.3}}
\psfrag{alpha0.4}[l]{\small{$\alpha_0$:0.4}}
\psfrag{alpha0.5}[l]{\small{$\alpha_0$:0.5}}
\psfrag{alpha0.6}[l]{\small{$\alpha_0$:0.6}}
\psfrag{alpha0.7}[l]{\small{$\alpha_0$:0.7}}
\psfrag{alpha0.8}[l]{\small{$\alpha_0$:0.8}}
\psfrag{alpha0.9}[l]{\small{$\alpha_0$:0.9}}
\includegraphics[width=0.49\textwidth]{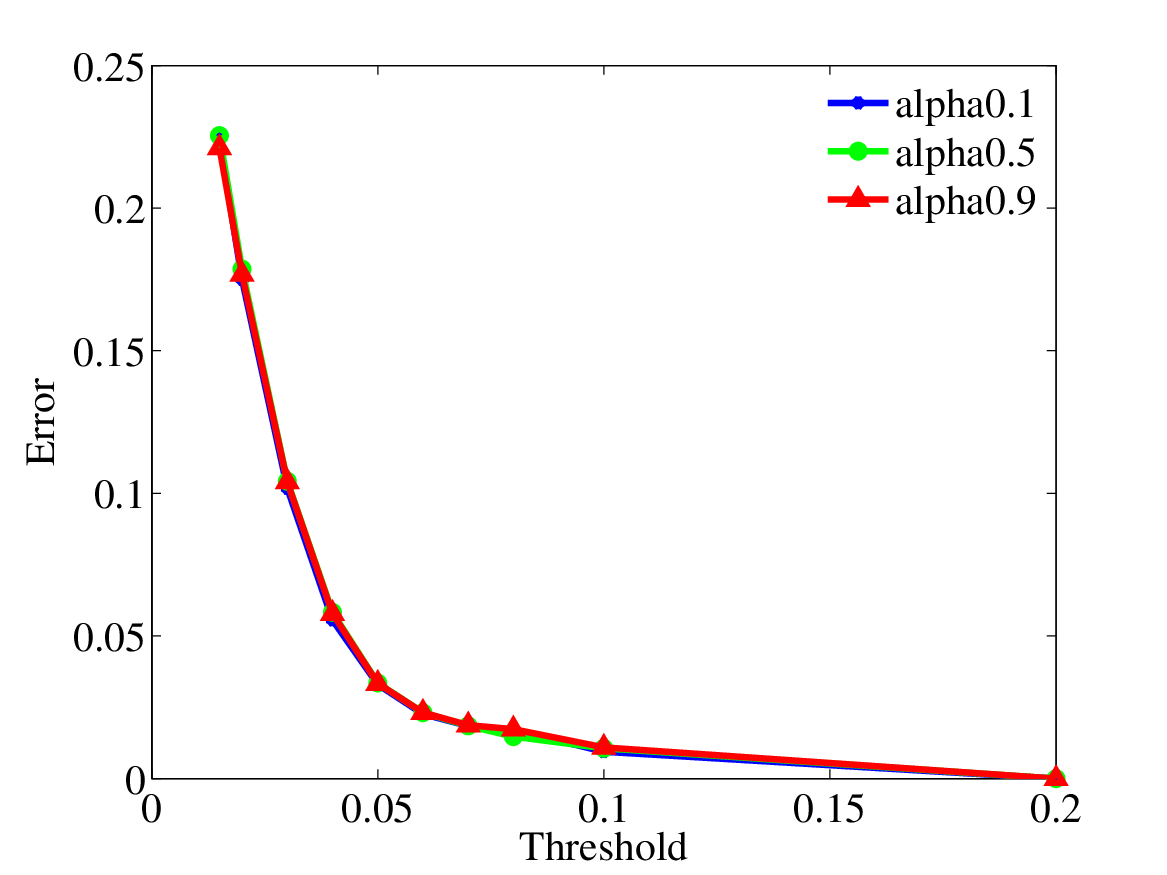}
\caption{Performance analysis of Facebook dataset under different settings of the concentration parameter ($\alpha_0$) for $\hat{k}=100$. 
}
\label{Fig:alpha0s}
\end{figure*}

We also obtain results for a range of values of $\alpha_0$ (Figure~\ref{Fig:alpha0s}). We observe that the recovery ratio improves with larger $\alpha_0$ since a larger $\alpha_0$ can recover overlapping communities more efficiently  while the error score remains relatively the same.


For the Facebook dataset, the top ten communities recovered with  lowest error consist of certain high schools, second majors and dorms/houses.
We observe that high school attributes are easiest to recover and second major and dorm/house are reasonably easy to recover by looking  at the friendship relations in Facebook.  This is reasonable:  college students from the same high school have a high probability of being friends; so do colleges students from the same dorm. 

\paragraph{DBLP Dataset: }

The DBLP data contains bibliographic records\footnote{\url{http://dblp.uni-trier.de/xml/Dblp.xml} } with various publication venues, such as journals and conferences, which we model as communities. We then consider authors who have published at least one paper in a community (publication venue) as a member of it. Co-authorship is thus modeled as link in the graph in which authors are represented as nodes.  In this framework, we could recover the top authors in communities and bridging authors.

\section{Conclusion}
In this paper, we presented a fast and unified moment-based framework for learning overlapping communities as well as topics in a corpus. There are several key insights involved. Firstly, our approach follows from a systematic and guaranteed learning procedure in contrast to several heuristic approaches which may not have strong statistical recovery guarantees. Secondly, though using a moment-based formulation may seem computationally expensive at first sight, implementing implicit ``tensor'' operations leads to significant speed-ups of the algorithm. Thirdly, employing randomized methods for spectral methods is promising in the computational domain, since the running time can then be significantly reduced.

This paper paves the way for several interesting directions for further research. While our current deployment incorporates community detection in a single graph, extensions to multi-graphs and hypergraphs are possible in principle. A careful and efficient implementation for such settings will be useful in a number of applications.  It is natural to extend the deployment to even larger datasets by having cloud-based systems. The issue of efficient partitioning of data and reducing communication between the machines becomes significant there. Combining our approach with other simple community detection approaches to gain even more speedups can be explored.

\section*{Acknowledgement}
The first author is supported by NSF BIGDATA IIS-1251267, the second author is supported in part by UCI graduate fellowship and NSF Award CCF-1219234, and the last author is supported in part by Microsoft Faculty Fellowship,  NSF Career award CCF-1254106, NSF Award CCF-1219234,  and ARO YIP Award W911NF-13-1-0084. The authors acknowledge insightful discussions with Prem Gopalan, David Mimno, David Blei, Qirong Ho, Eric Xing, Carter Butts, Blake Foster, Rui Wang, Sridhar Mahadevan, and the CULA team. Special thanks to Prem Gopalan and David Mimno for providing the variational code and answering all our questions. The authors also thank Daniel Hsu and Sham Kakade for initial discussions regarding the implementation of the tensor method. We also thank Dan Melzer for helping us with the system-related issues.

\bibliography{community}

\section{Appendix}
\appendix

\section{Stochastic Updates}
\label{sec:apdx_update}
After obtaining the whitening matrix, we whiten the data $G^\top_{x,A}$, $G^\top_{x,B}$ and $G^\top_{x,C}$ by linear operations to get $y^t_A$, $y^t_B$ and $y^t_C\in \mathbb{R}^{k}$:
\begin{align*}
y^t_A : = \left<G^\top_{x,A}, W \right>,
\; 
y^t_B := \left<Z_B G^\top_{x,B},W \right>,
\; 
y^t_C& : = \left<Z_C G^\top_{x,C},W \right>.
\end{align*}
where $x\in X$ and $t$ denotes the index of the online data.

The stochastic gradient descent algorithm is obtained by taking the derivative of the loss function $\frac{\partial L^t(\mathbf{v})}{\partial v_i}$:
\begin{align*}
\frac{\partial L^t(\mathbf{v})}{\partial v_i}=&
\theta\sum\limits_{j=1}^{k} \left<v_j,v_i\right>^2 v_j
- \frac{(\alpha_0+1)(\alpha_0+2)}{2} \left<v_i, y_A^t\right> \left<v_i, y_B^t\right> y_C^t
- \alpha_0^2 \left<\phi_i^t,\bar{y}_A\right>\left<\phi_i^t,\bar{y}_B^t\right>\bar{y}_C \\
&+ \frac{\alpha_0(\alpha_0+1)}{2}\left<\phi_i^t, y_A^t\right>\left<\phi_i^t, y_B^t\right>\bar{y}_C
+\frac{\alpha_0(\alpha_0+1)}{2}\left<\phi_i^t, y_A^t\right>\left<\phi_i^t, \bar{y}_B\right>y_C
+\frac{\alpha_0(\alpha_0+1)}{2}\left<\phi_i^t,\bar{y}_A\right>\left<\phi_i^t,y_B^t\right>y_C
\end{align*}
for $i \in [k]$, where $y_A^t$, $y_B^t$ and $y_C^t$ are the online whitened data points as discussed in the whitening step and $\theta$ is a constant factor that we can set.

The iterative updating equation for the stochastic gradient update is given by
\begin{equation}
\phi_i^{t+1} \leftarrow \phi_i^t - \beta^t \frac{\partial L^t}{\partial v_i}\llvert_{\phi_i^t}
\end{equation}
for $i \in [k]$, where $\beta^t$ is the learning rate, $\phi^t_i$ is the last iteration eigenvector and $\phi^t_i$ is the updated eigenvector.
We update eigenvectors through
\begin{align}
\phi_i^{t+1} \leftarrow  \phi_i^t  & - \theta\beta^t \sum\limits_{j=1}^{k} \left[\left<\phi_j^t,\phi_i^t\right>^2 \phi_j^t\right]
+ \text{shift} [ \beta^t \left<\phi_i^t, y_A^t\right> \left<\phi_i^t, y_B^t\right> y_C^t ]\label{eq:term2_1}
\end{align}

Now we shift the updating steps so that they correspond to the centered Dirichlet moment forms, i.e.,
\begin{align}
& \text{shift}[ \beta^t \left<\phi_i^t, y_A^t\right> \left<\phi_i^t, y_B^t\right> y_C^t ]
 :=
\beta^t \frac{(\alpha_0+1)(\alpha_0+2)}{2} \left<\phi_i^t,y_A^t\right> \left<\phi_i^t, y_B^t\right> y_C^t 
+  \beta^t {\alpha_0^2}\left<\phi_i^t,\bar{y}_A\right> \left<\phi_i^t, \bar{y}_B\right>\bar{y}_C \nonumber 
\\
&- \beta^t \frac{\alpha_0(\alpha_0+1)}{2}\left<\phi_i^t, y_A^t\right>\left<\phi_i^t, y_B^t\right>\bar{y}_C 
-  \beta^t \frac{\alpha_0(\alpha_0+1)}{2}\left<\phi_i^t, y_A^t\right>\left<\phi_i^t, \bar{y}_B\right>y_C 
-  \beta^t \frac{\alpha_0(\alpha_0+1)}{2}\left<\phi_i^t,\bar{y}_A\right>\left<\phi_i^t,y_B^t\right>y_C,
\end{align}
where $\bar{y}_A:= \mathbb{E}_t [y_A^t]$ and similarly for $\bar{y}_B$ and $\bar{y}_C$.

\section{Proof of correctness of our algorithm:}
We now prove the correctness of our algorithm.

First, we compute $M_2$ as just $$\mathbb{E}_x\left[ \tilde{G}_{x,C}^\top \otimes \tilde{G}_{x,B}^\top| \Pi_A, \Pi_B, \Pi_C\right]$$
where we define
\begin{align*}
\tilde{G}_{x,B}^\top & := \mathbb{E}_x\left[G_{x,A}^\top \otimes G_{x,C}^\top\llvert \Pi_A, \Pi_C\right] \left(\mathbb{E}_x\left[G_{x,B}^\top \otimes G_{x,C}^\top \llvert \Pi_B, \Pi_C\right]\right)^\dag G_{x,B}^\top
\\
\tilde{G}_{x,C}^\top & :=\mathbb{E}_x\left[G_{x,A}^\top \otimes G_{x,B}^\top\llvert \Pi_A, \Pi_B \right]\left(\mathbb{E}_x\left[G_{x,C}^\top \otimes G_{x,B}^\top\llvert \Pi_B, \Pi_C \right]\right)^{\dagger} G_{x,C}^\top.
\end{align*}
Define $F_A$ is defined as $F_A := \Pi_A^\top P^\top$, we obtain $M_2=\mathbb{E}\left[G^\top_{x,A} \otimes G^\top_{x,A}\right] = \Pi_A^\top P^\top \left(\mathbb{E}_x[\pi_x \pi_x^\top]\right) P \Pi_A = F_A\left(\mathbb{E}_x[\pi_x \pi_x^\top]\right)F_A^\top$. Note that $P$ is the community connectivity matrix defined as $P\in [0,1]^{k\times k}$.
Now that we know $M_2$, $\mathbb{E}\left[\pi_i^2\right]= \frac{\alpha_i(\alpha_i+1)}{\alpha_0(\alpha_0+1)}$, and $\mathbb{E}\left[\pi_i \pi_j\right]= \frac{\alpha_i \alpha_j}{\alpha_0(\alpha_0+1)} \forall i\neq j$, we can get the centered second order moments $\Pairs^{\community}$ as

\begin{align}
\Pairs^{\community} & := F_A \text{ diag}\left(\left[\frac{\alpha_1 \alpha_1+1}{\alpha_0(\alpha_0+1)},\ldots,\frac{\alpha_k \alpha_k+1}{\alpha_0(\alpha_0+1)}\right]\right)F_A^\top
\\
& = M_2 - \frac{\alpha_0}{\alpha_0+1} F_A \left(\hat{\alpha} \hat{\alpha}^\top - \text{ diag}\left(\hat{\alpha} \hat{\alpha}^\top	 \right)\right)F_A^\top
\\
& = \frac{1}{n_X} \sum\limits_{x\in X} Z_C G_{x,C}^\top G_{x,B} Z_B^\top  -\frac{\alpha_0}{\alpha_0+1} \left(\mu_{A} \mu_{A}^\top- \text{ diag}\left(\mu_{A} \mu_{X\rightarrow A}^\top\right) \right)
\end{align}
Thus, our whitening matrix is computed. Now, our whitened tensor is $\mathcal{T} $ is given by
\begin{align*}
\mathcal{T} & = \mathcal{T}^{\community} (W,W,W) = \frac{1}{n_X} \sum_{x}\left[(W^\top F_A \pi^{\alpha_0}_x) \otimes (W^\top F_A \pi^{\alpha_0}_x) \otimes (W^\top F_A \pi^{\alpha_0}_x)\right],
\end{align*}
where $\pi^{\alpha_0}_x$ is the centered vector so that $\Ebb[\pi_x^{\alpha_0}\otimes \pi_x^{\alpha_0}\otimes \pi_x^{\alpha_0}]$ is diagonal.
We then apply the stochastic gradient descent technique to decompose the third order moment.

\section{GPU Architecture}
\label{sec:apdx_arch}

The algorithm we propose is very amenable to parallelization and is scalable which makes it suitable to implement on processors with multiple cores in it. Our method consists of simple linear algebraic operations, thus enabling us to utilize \emph{Basic Linear Algebra Subprograms} (BLAS) routines such as BLAS I (vector operations), BLAS II (matrix-vector operations), BLAS III (matrix-matrix operations), Singular Value Decomposition (SVD),  and iterative operations such as stochastic gradient descent for tensor decomposition that can easily take advantage of Single Instruction Multiple Data (SIMD)  hardware units present in the GPUs. As such, our method is amenable to parallelization and is  ideal for GPU-based implementation.

\paragraph{Overview of code design: }From a higher level point of view, a typical GPU based computation is a three step process involving data transfer from CPU memory to GPU global memory, operations on the data now present in GPU memory and finally, the result transfer from the GPU memory back to the CPU memory. We use the CULA library for implementing the linear algebraic operations. 

\paragraph{GPU compute architecture: } The GPUs achieve massive parallelism by having hundreds of homogeneous processing cores integrated on-chip. Massive replication of these cores provides the parallelism needed by the applications that run on the GPUs. These cores, for the Nvidia GPUs, are known as \emph{CUDA cores}, where each core has fully pipelined floating-point and integer arithmetic logic units. In Nvidia's Kepler architecture based GPUs
, these CUDA cores are bunched together to form a \emph{Streaming Multiprocessor} (SMX). These SMX units act as the basic building block for Nvidia Kepler GPUs. Each GPU contains multiple SMX units where each SMX unit has 192 single-precision CUDA cores, 64 double-precision units, 32 special function units, and 32 load/store units for data movement between cores and memory.

Each SMX has L$1$, shared memory and a read-only data cache that are common to all the CUDA cores in that SMX unit. Moreover, the programmer can choose between different configurations of the shared memory and L$1$ cache. Kepler GPUs also have an L$2$ cache memory of about $1.5$MB that is common to all the on-chip SMXs. Apart from the above mentioned memories, Kepler based GPU cards come with a large DRAM memory, also known as the global memory, whose size is usually in gigabytes. This global memory is also visible to all the cores. The GPU cards usually do not exist as standalone devices. Rather they are part of a CPU based system, where the CPU and GPU interact with each other via PCI (or PCI Express) bus.

In order to program these massively parallel GPUs, Nvidia provides a framework known as \emph{CUDA} that enables the developers to write programs in languages like C, C++, and Fortran etc. A CUDA program constitutes of functions called \emph{CUDA kernels} that execute across many parallel software threads, where each thread runs on a CUDA core. Thus the GPU's performance and scalability is exploited by the simple partitioning of the algorithm into fixed sized blocks of parallel threads that run on hundreds of CUDA cores. The threads running on an SMX can synchronize and cooperate with each other via the shared memory of that SMX unit and can access the Global memory. Note that the CUDA kernels are launched by the CPU but they get executed on the GPU. Thus compute architecture of the GPU requires CPU to initiate the CUDA kernels.

CUDA enables the programming of Nvidia GPUs by exposing low level API. Apart from CUDA framework, Nvidia provides a wide variety of other tools and also supports third party libraries that can be used to program Nvidia GPUs. Since a major chunk of the scientific computing algorithms is linear algebra based, it is not surprising that the standard linear algebraic solver libraries like BLAS and \emph{Linear Algebra PACKage} (LAPACK) also have their equivalents for Nvidia GPUs in one form or another. Unlike CUDA APIs, such libraries expose APIs at a much higher-level and mask the architectural details of the underlying GPU hardware to some extent thus enabling relatively faster development time.

Considering the tradeoffs between the algorithm's computational requirements, design flexibility, execution speed and development time, we choose \emph{CULA-Dense} as our main implementation library. CULA-Dense provides GPU based implementations of the LAPACK and BLAS libraries for dense linear algebra and contains routines for systems solvers, singular value decompositions, and eigen-problems. Along with the rich set of functions that it offers, CULA provides the flexibility needed by the programmer to rapidly implement the algorithm while maintaining the performance. It hides most of the GPU architecture dependent programming details thus making it possible for rapid prototyping of GPU intensive routines.

\begin{figure}[H]
	\centering
	\psfrag{Logarithm of the buffer size divided by 8}[l]{\scriptsize{$\log\left(\frac{\text{buffer size}}{8}\right)$}}
	\psfrag{Time (in seconds)}[c]{\scriptsize{Time (s)}}
	\psfrag{CPU-GPU buffer round-trip truncation time}[c]{}
	\includegraphics[width=0.4\textwidth]{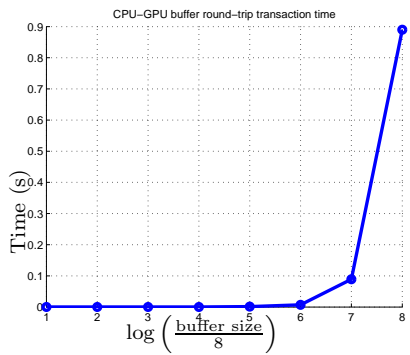}
	\caption{Experimentally measured time taken for buffer transfer between the CPU and the GPU memory in our system.}
	\label{tran_time_expt}
\end{figure}

The data transfers between the CPU memory and the GPU memory are usually explicitly initiated by CPU and are carried out via the PCI (or PCI Express) bus interconnecting the CPU and the GPU. The movement of data buffers between CPU and GPU is the most taxing in terms of time. The buffer transaction time is shown in the plot in Figure~\ref{tran_time_expt}. Newer GPUs, like Kepler based GPUs, also support useful features like GPU-GPU direct data transfers without CPU intervention.
Our system and software specifications are given in Table~\ref{tab:sys_spec}.

CULA exposes two important interfaces for GPU programming namely, \emph{standard} and \emph{device}.
Using the standard interface, the developer can program without worrying about the underlying architectural details of the GPU as the standard interface takes care of all the data movements, memory allocations in the GPU and synchronization issues. This however comes at a cost.
For every standard interface function call the data is moved in and out of the GPU even if the output result of one operation is directly required by the subsequent operation.
This unnecessary movement of intermediate data can dramatically impact the performance of the program.
In order to avoid this, CULA provides the device interface.
We  use the  device interface for STGD in which the programmer is responsible for data buffer allocations in the GPU memory, the required data movements between the CPU and GPU, and operates only on the data in the GPU. Thus the subroutines of the program that are iterative in nature are good candidates for device implementation. 

\paragraph{Pre-processing and post-processing: }
The pre-processing involves matrices whose leading dimension is of the order of number of nodes. These are implemented using the CULA standard interface BLAS II and BLAS III routines.

Pre-processing requires SVD computations for the Moore-Penrose pseudoinverse calculations. We use CULA SVD routines since  these SVD operations are carried out on matrices of moderate size.
We further replaced the CULA SVD routines with more scalable SVD and pseudo inverse routines using random projections~\citep{gittens2013revisiting} to handle larger datasets such as DBLP dataset in our experiment.

After STGD, the community membership matrix estimates are obtained using BLAS III routines provided by the CULA standard interface. The matrices are then used for hypothesis testing to evaluate the algorithm against the ground truth.

\section{Results on Synthetic Datasets}
\label{sec:apdx_synth}
{\em Homophily} is an important factor in social interactions~\citep{mcpherson2001birds}; the term {\em homophily} refers to the tendency that actors in the same community interact more than across different communities. Therefore, we assume diagonal dominated community connectivity matrix $P$ with diagonal elements equal to $0.9$ and off-diagonal elements equal to $0.1$.
Note that $P$ need neither be stochastic nor symmetric. Our algorithm allows for randomly generated community connectivity matrix $P$ with support $[0,1]$. In this way, we look at general directed social ties among communities.

\begin{table}
   \centering
   \begin{tabular}{@{} llllr @{}}
      \toprule
      $n$  & $k$ & $\alpha_0$ &  Error& Time (secs)\\
      \midrule
       1e2 	&10	&	0	& 0.1200	   & 0.5\\
       1e3 	&10	&	0	&  0.1010  & 1.2\\
       1e4 	&10	&	0 	&   0.0841 & 43.2\\
       1e2 	&10	&	1	&   0.1455	& 0.5\\
       1e3 	&10	&	1	&   0.1452	& 1.2\\
       1e4 	&10	&	1 	&   0.1259	& 42.2\\
      \bottomrule
   \end{tabular}
   \caption{Synthetic simulation results for different configurations. Running time is the time taken to run to convergence.}
   \label{tab:synResult}
\end{table}

We perform experiments for both the stochastic block model ($\alpha_0=0$) and the mixed membership model. For the mixed membership model, we set the concentration parameter $\alpha_0 =1$. We note that the error is around $8\% - 14\%$ and the running times are under a minute, when $n \leq 10000$ and $n \gg k$.

The results are given in Table~\ref{tab:synResult}.
We observe that more samples result in a more accurate recovery of memberships which matches intuition and theory. Overall, our learning algorithm performs better in the stochastic block model case than in the mixed membership model case although we note that the accuracy is quite high for practical purposes. Theoretically, this is expected since smaller concentration parameter $\alpha_0$ is easier for our algorithm to learn~\citep{AnandkumarEtal:community12COLT}. Also, our algorithm is scalable to an order of magnitude more in $n$ as illustrated by experiments on real-world large-scale datasets.

\section{Comparison of Error Scores}
\label{sec:otherscores} 
Normalized Mutual Information (NMI) score~\citep{lancichinetti2009detecting} is another popular score which is defined differently for overlapping and non-overlapping community models. For non-overlapping block model, ground truth membership for node $i$ is a discrete $k$-state categorical variable $\Pi_{\text{block}} \in [k]$ and the estimated membership is a discrete $\widehat{k}$-state categorical variable $\widehat{\Pi}_{\text{block}} \in [\widehat{k}]$. 
The empirical distribution of ground truth membership categorical variable  $\Pi_{\text{block}}$ is easy to obtain. Similarly is the empirical distribution of the estimated membership categorical variable $\widehat{\Pi}_{\text{block}}$.  NMI for block model is defined as
\begin{align*}
& N_{\text{block}}(\widehat{\Pi}_{\text{block}}:\Pi_{\text{block}}) :=  \frac{H(\Pi_{\text{block}})+H(\widehat{\Pi}_{\text{block}})-H(\Pi_{\text{block}},\widehat{\Pi}_{\text{block}})}{\left(H(\Pi_{\text{block}})+H(\widehat{\Pi}_{\text{block}})\right)/2} .
\end{align*}

The NMI for overlapping communities is  a binary vector instead of a categorical variable~\citep{lancichinetti2009detecting}.  The ground truth membership for node $i$ is a binary vector of length $k$, $\mathbf{\Pi}_{{\text{mix}}}$, while the estimated membership for node $i$ is a binary vector of length $\widehat{k}$, $\mathbf{\widehat{\Pi}}_{{\text{mix}}}$. This notion coincides with one column of our membership matrices $\Pi\in\mathbb{R}^{k\times n}$ and $\widehat{\Pi}\in \mathbb{R}^{\widehat{k} \times n}$ except that our membership matrices are stochastic. In other words, we consider all the nonzero entries of $\Pi$ as 1's, then each column of our $\Pi$ is a sample for $\Pi_{{\text{mix}}}$. The $m$-th entry of this binary vector is the realization of a random variable $\Pi_{{\text{mix}}_m} = (\mathbf{\Pi}_{{\text{mix}}})_m$, whose probability distribution is
\[
P(\Pi_{{\text{mix}}_m}=1)= \frac{n_m}{n}, \quad P(\Pi_{{\text{mix}}_m}=0)= 1-\frac{n_m}{n},
\]
where $n_m$ is the number of nodes in community $m$. The same holds for $\widehat{\Pi}_{\text{mix}_m}$. The normalized conditional entropy between $\mathbf{\Pi}_{{\text{mix}}}$ and $\mathbf{\widehat{\Pi}}_{{\text{mix}}}$ is defined as
\begin{equation}\label{eqn:nmi}
H(\mathbf{\widehat{\Pi}}_{{\text{mix}}} \lvert \mathbf{{\Pi}}_{{\text{mix}}})_{\text{norm}}  :=\frac{1}{k} \sum_{j \in [k]} \min_{i \in [\widehat{k}]}
\frac{H\left(\widehat{\Pi}_{{\text{mix}}_i}  \lvert  \Pi_{{\text{mix}}_j} \right)}{H(\Pi_{{\text{mix}}_j} )}
\end{equation}
where $\Pi_{\text{mix}_j}$ denotes the $j^{th}$ entry of $\mathbf{\Pi}_{\text{mix}}$ and similarly for $\widehat{\Pi}_{\text{mix}_i}$.
The NMI for overlapping community is
\begin{align*}
& N_{\text{mix}}(\mathbf{\widehat{\Pi}}_{\text{mix}} : \mathbf{\Pi} _{\text{mix}}):= 1-\frac{1}{2}\left[H(\mathbf{\Pi}_{\text{mix}} \lvert \mathbf{\widehat{\Pi}}_{\text{mix}})_{\text{norm}}+H(\mathbf{\widehat{\Pi}}_{\text{mix}} \lvert \mathbf{\Pi}_{\text{mix}})_{\text{norm}}\right].
\end{align*}

\begin{figure*}[h]
   \centering
   \psfrag{dense pi1}[r]{dense $\Pi_1$ }
   \psfrag{sparse pi2}[r]{sparse $\Pi_2$ }
   \psfrag{length n vector}[c]{length $n$ membership vector}
   \psfrag{zeros}[l]{$0$}
   \psfrag{ones}[l]{$1$}
   \psfrag{dense community}[l]{large sized community}
   \psfrag{sparse community}[l]{small sized community}
   \includegraphics[width=0.7\textwidth]{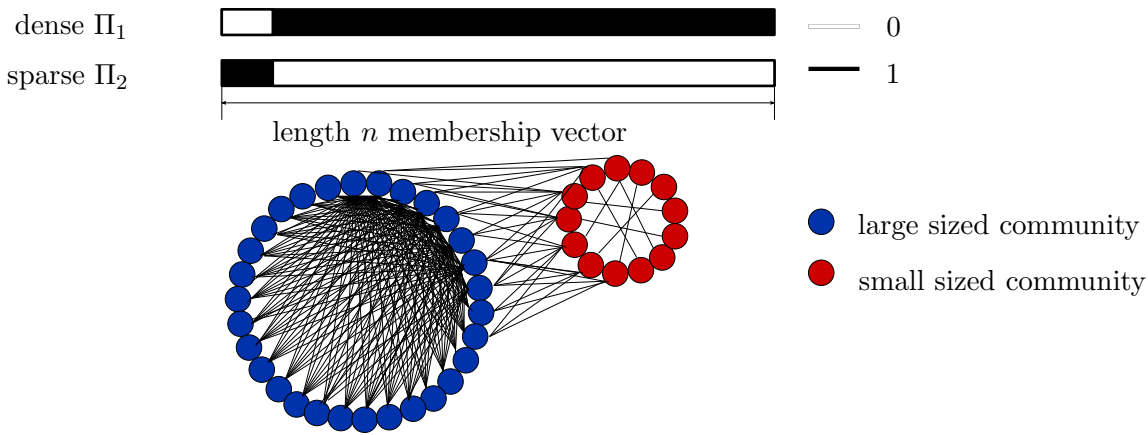}
   \caption{ A special case of a pair of extremely dense and sparse communities. Theoretically, the sparse community is more difficult to recover than the dense one. However, the NMI score penalizes both of them equally. Note that for dense $\Pi_1$, $P(\Pi_{\text{mix}_1}=0)=\frac{\text{\# of 0s in } \Pi_1}{n}$ which is equal to $P(\Pi_{\text{mix}_2}=1)=\frac{\text{\# of 1s in } \Pi_2}{n}$. Similarly, $P(\Pi_{\text{mix}_1}=1)=\frac{\text{\# of 1s in } \Pi_1}{n}$ which is equal to $P(\Pi_{\text{mix}_2}=0)=\frac{\text{\# of 0s in } \Pi_2}{n}$. Therefore, $H(\Pi_{\text{mix}_1})=H(\Pi_{\text{mix}_2})$.}
      \label{fig:NMI}
\end{figure*}

There are two aspects in evaluating the error.
The first aspect is the $l_1$ norm error. According to Equation~\eqref{eqn:nmi},  the error function used in NMI score is  $\frac{H\left(\widehat{\Pi}_{{\text{mix}}_i}  \lvert  \Pi_{{\text{mix}}_j} \right)}{H(\Pi_{{\text{mix}}_j} )}$.
NMI is not suitable for evaluating recovery of different sized communities. In the special case of a pair of  extremely sparse and dense membership vectors, depicted in Figure~\ref{fig:NMI}, ${H(\Pi_{\text{mix}_j})}$ is the same for both the dense and the sparse vectors since they are flipped versions of each other (0s flipped to 1s and vice versa).  However, the smaller sized community (i.e. the sparser community vector),  shown in red in Figure~\ref{fig:NMI}, is significantly more difficult to recover than the larger sized community shown in blue in Figure~\ref{fig:NMI}. Although this example is an extreme scenario that is not seen in practice, it justifies the drawbacks of the NMI. Thus, NMI is not suitable for evaluating recovery of different sized communities.
In contrast,   our error function employs a normalized $l_1$ norm error which penalizes more for larger sized communities than smaller ones.

The second aspect is the error induced by false pairings of estimated and ground-truth communities. NMI score selects only the closest estimated community through normalized conditional entropy minimization and it does not account for statistically significant dependence between an estimated community and multiple ground truth communities and vice-versa, and therefore it underestimates error.  However, our error score does not limit to a matching between the estimated and ground truth communities:  if an estimated community is found to have statistically significant correlation with multiple ground truth communities (as evaluated by the $p$-value), we penalize for the error over all such ground truth communities. Thus, our error score is a harsher measure of evaluation than NMI.
This notion of  ``soft-matching'' between ground-truth and estimated communities also enables validation of recovery of a combinatorial union of communities instead of single ones.

A number of other scores such as ``separability'', ``density'', ``cohesiveness'' and ``clustering coefficient''~\citep{yang2012defining} are non-statistical measures of faithful community recovery. 
 The scores of~\citep{yang2012defining}  intrinsically aim to evaluate the level of clustering within a community.  However our goal is to measure the accuracy of recovery of the communities and not how well-clustered the communities are. 

Banerjee and Langford~\citep{banerjee2004objective} proposed an objective evaluation criterion for clustering which use classification performance as the evaluation measure. In contrast, we look at how well the method performs in recovering the hidden communities, and we are not evaluating predictive performance. Therefore, this measure is not used in our evaluation.

Finally, we note that cophenetic correlation is another statistical score used for evaluating clustering methods, but note that it is only valid for hierarchical clustering and it is a measure of how faithfully a dendrogram preserves the pairwise distances between the original unmodeled data points~\citep{sokal1962comparison}. Hence, it is not employed in this paper.

\end{document}